\documentclass[5p, twocolumn]{elsarticle} 
\usepackage{amsmath,amssymb,amsfonts}
\usepackage[ruled]{algorithm2e}
\usepackage{booktabs, colortbl, makecell}
\usepackage{multirow}
\usepackage{stfloats}
\usepackage{xcolor}

\usepackage[colorlinks, linkcolor=blue, anchorcolor=blue, citecolor=blue]{hyperref}
\usepackage{cleveref}
\usepackage[normalem]{ulem}

\definecolor{color_F2F2F2}{HTML}{F2F2F2}

\newdefinition{remark}{Remark}
\newdefinition{definition}{Definition}
\newdefinition{property}{Property}
\newproof{proof}{Proof} 

\graphicspath{{Figures/PDF/}} % default path of Figures.

\journal{Information Fusion}

\begin{document}
\begin{frontmatter}
	\title{PhyDAE: Physics-Guided Degradation-Adaptive Experts for All-in-One Remote Sensing Image Restoration}
	
	\author[hit,sirs]{Zhe Dong}
	% \ead{wqzhao@stu.hit.edu.cn}
	
	\author[hit,sirs]{Yuzhe Sun}
	
	\author[hit,sirs]{Haochen Jiang}
	
	\author[hit,sirs]{Tianzhu Liu}
	% \ead{tzliu@hit.edu.cn}
	
	\author[hit,sirs]{Yanfeng Gu\corref{cor1}}
	\ead{guyf@hit.edu.cn}
	
	\cortext[cor1]{Corresponding author}
	
	\affiliation[hit]{
		organization={School of Electronic and Information Engineering, Harbin Institute of Technology}, 
		addressline={Xidazhi Street}, 
		city={Harbin},
		postcode={150001}, 
		state={Heilongjiang}, 
		country={China}
	}

	\affiliation[sirs]{
		organization={Heilongjiang Province Key Laboratory of Space-Air-Ground Integrated Intelligent Remote Sensing}, 
		addressline={Yikuang Street}, 
		city={Harbin},
		postcode={150001}, 
		state={Heilongjiang}, 
		country={China}
	}

	\begin{abstract}
		Remote sensing images inevitably suffer from various degradation factors during acquisition, including atmospheric interference, sensor limitations, and imaging conditions. These complex and heterogeneous degradations pose severe challenges to image quality and downstream interpretation tasks. Addressing limitations of existing all-in-one restoration methods that overly rely on implicit feature representations and lack explicit modeling of degradation physics, this paper proposes Physics-Guided Degradation-Adaptive Experts (PhyDAE). The method employs a two-stage cascaded architecture transforming degradation information from implicit features into explicit decision signals, enabling precise identification and differentiated processing of multiple heterogeneous degradations including haze, noise, blur, and low-light conditions. The model incorporates progressive degradation mining and exploitation mechanisms, where the Residual Manifold Projector (RMP) and Frequency-Aware Degradation Decomposer (FADD) comprehensively analyze degradation characteristics from manifold geometry and frequency perspectives. Physics-aware expert modules and temperature-controlled sparse activation strategies are introduced to enhance computational efficiency while ensuring imaging physics consistency. Extensive experiments on three benchmark datasets (MD-RSID, MD-RRSHID, and MDRS-Landsat) demonstrate that PhyDAE achieves superior performance across all four restoration tasks, comprehensively outperforming state-of-the-art methods. Notably, PhyDAE substantially improves restoration quality while achieving significant reductions in parameter count and computational complexity, resulting in remarkable efficiency gains compared to mainstream approaches and achieving optimal balance between performance and efficiency. Code is available at https://github.com/HIT-SIRS/PhyDAE.
	\end{abstract}
	
\end{frontmatter}

\section{Introduction}

Remote sensing imagery has emerged as an indispensable technological means for Earth observation, environmental monitoring, disaster assessment, and urban planning. With its large-scale spatial coverage and temporal continuity, remote sensing plays an irreplaceable role in global change research and fine-grained management. However, remote sensing images inevitably suffer from multiple degradation factors during acquisition, including atmospheric interference, sensor limitations, and imaging conditions\citep{xiao2023ediffsr}. Typical degradations include haze pollution caused by atmospheric scattering, sensor electronic noise, motion blur induced by platform movement, and low-light degradation under weak illumination conditions. These complex and heterogeneous degradations often coexist in superimposed forms and exhibit high spatial and intensity heterogeneity, significantly undermining image visualization quality and downstream intelligent interpretation performance, thereby posing fundamental challenges to automated remote sensing applications.

Traditional image restoration methods primarily comprise filter-based signal processing techniques and optimization-based variational approaches\citep{he2010single}. Although theoretically well-established, these methods often rely on simplified assumptions when handling complex degradation patterns, making it difficult to effectively model the nonlinear characteristics of real-world degradations. With the advancement of deep learning technologies, convolutional neural network-based restoration methods have demonstrated remarkable advantages through their powerful capability to model complex degradation patterns\citep{liang2021swinir}. However, early approaches predominantly designed specialized networks for individual degradations. While achieving satisfactory performance in respective tasks, cascading single-task models in multi-degradation remote sensing scenarios incurs prohibitive computational costs and fails to effectively leverage shared characteristics across degradations. 

Recent research has progressively focused on all-in-one image restoration frameworks\citep{li2022all}, attempting to simultaneously address multiple degradation types within a unified architecture through parameter-efficient strategies such as dynamic convolution, prompt learning, or mixture-of-experts mechanisms to achieve task adaptation. Nevertheless, these methods still encounter three critical bottlenecks in remote sensing applications: First, excessive reliance on implicit feature representations without explicit modeling of degradation physical characteristics, limiting their capability to handle complex atmospheric and sensor-related degradations. Second, absence of progressive mechanisms for mining and exploiting degradation patterns, where direct end-to-end mapping struggles to accommodate the high variability and compound nature of remote sensing degradations. Third, lack of physics-based constraint guidance, resulting in restoration outcomes that, despite visual plausibility, may violate imaging physics principles and consequently compromise the reliability of subsequent quantitative analysis.

The fundamental challenge in remote sensing image restoration lies in the multi-scale coupling characteristics of complex degradations\citep{zhang2022single}. Unlike natural images, remote sensing imaging involves multiple physical processes including atmospheric propagation and sensor response\citep{cheng2023restoration}, forming highly nonlinear and cross-scale coupled degradation manifolds. These degradation factors exhibit strong spatial heterogeneity and significant dynamic variations in degradation intensity, rendering traditional restoration paradigms based on natural image assumptions inadequate\citep{zhu2024mwformer}. Consequently, there is an urgent need for novel restoration frameworks that integrate physical priors and adapt to multi-source degradations, enabling precise modeling and efficient restoration of remote sensing image degradations.

Addressing the aforementioned challenges, this paper proposes Physics-Guided Degradation-Adaptive Experts (PhyDAE) for all-in-one remote sensing image restoration. The method employs a two-stage cascaded architecture to transform degradation information from implicit features into explicit decision signals, thereby establishing adaptive mapping between degradation patterns and restoration strategies, capable of differentiated processing of multiple heterogeneous degradations including haze, noise, blur, and low-light conditions. Building upon this foundation, the model further introduces a progressive degradation mining and exploitation mechanism that first extracts degradation residual information through unconditional restoration, then encodes it as multi-scale conditional signals to guide precise restoration, achieving layer-wise transition from degradation discovery to degradation adaptation while leveraging complementary advantages of frequency-domain and spatial-domain features to effectively characterize and eliminate complex compound degradations. Simultaneously, PhyDAE incorporates physics-aware expert modules within the mixture-of-experts architecture and implements dynamic routing through degradation-aware frequency-domain embeddings, enabling the model to achieve task discriminability and efficient processing while adhering to imaging physics constraints, thereby significantly enhancing restoration quality and model interpretability.

The main contributions of this paper are summarized as follows:
\begin{itemize}
	\item We propose an all-in-one remote sensing restoration framework that deeply integrates degradation physics priors with mixture-of-experts mechanisms, revealing the intrinsic connections among degradation modeling physicality, restoration decision adaptability, and model interpretability.
	\item We innovatively design a progressive degradation mining and exploitation mechanism that achieves hierarchical modeling from degradation discovery to precise adaptation, effectively addressing modeling challenges in compound degradation scenarios.
	\item We construct physics-aware expert modules and dynamic routing mechanisms that enhance restoration accuracy and model interpretability while ensuring imaging physics consistency, providing a novel technical pathway for high-quality remote sensing image restoration.
\end{itemize}

\section{Related Works}
\subsection{All-in-One Image Restoration}

Traditional image restoration methods have predominantly focused on addressing specific degradation types, such as denoising\citep{zhang2017beyond}, deblurring\citep{kupyn2018deblurgan}, and dehazing\citep{qin2020ffa}, achieving remarkable performance within their respective domains. However, these task-specific approaches suffer from inherent limitations when confronted with mixed or unknown degradations commonly encountered in real-world scenarios. Early attempts at unified restoration, such as the parameterized image operator framework\citep{fan2019general}, laid the groundwork for handling multiple tasks within a single model. The seminal work by Li \textit{et al.}\citep{li2020all} introduced the first comprehensive all-in-one weather restoration framework, employing multi-encoder and decoder architectures with neural architecture search across task-specific components. While demonstrating the feasibility of unified processing, these pioneering approaches still required explicit degradation type identification and separate training protocols for different restoration tasks.

The development of truly task-agnostic architectures marked a significant paradigm shift in all-in-one image restoration research. TransWeather\citep{valanarasu2022transweather} pioneered transformer-based restoration using weather-type queries and learnable degradation representations, eliminating the need for explicit task specification. AirNet\citep{li2022all} further advanced this direction by introducing the first blind all-in-one image restoration approach, employing contrastive learning to extract degradation-aware features without prior knowledge of corruption types. Subsequent architectural innovations have explored various unified designs, including shared backbone architectures with multiple decoders\citep{han2022blind}, mixture-of-experts frameworks for adaptive expert selection\citep{yu2024multi}, and state-space models like MambaIR\citep{guo2024mambair} that leverage efficient sequence modeling for long-range dependency capture. These developments have collectively demonstrated that sophisticated feature disentanglement and adaptive routing mechanisms can enable effective multi-degradation handling within unified network architectures.

Contemporary all-in-one image restoration research has increasingly focused on advanced learning strategies that address the inherent challenges of multi-task optimization. Contrastive learning has proven particularly effective for degradation representation learning, with approaches like CPLPromptIR\citep{wu2025beyond} extending contrastive principles from feature-level discrimination to prompt-level alignment. The emergence of prompt-based learning has become the dominant paradigm, where PromptIR\citep{potlapalli2023promptir} pioneered the integration of learnable visual prompts as lightweight, adaptive modules for multi-scale degradation-aware guidance. This concept has evolved to encompass multimodal prompting, with TextPromptIR\citep{yan2025textual} and InstructIR\citep{conde2024instructir} demonstrating the effectiveness of natural language guidance for intuitive user interaction and enhanced generalization capabilities. Multi-task learning strategies have also matured, with works like GRIDS\citep{cao2024grids} introducing task grouping based on degradation correlations and TUR\citep{wu2025debiased} proposing task-aware optimization with adaptive regularization to mitigate conflicts between competing restoration objectives.

The integration of large-scale pretrained models represents the most recent advancement in all-in-one image restoration capabilities. Vision-language models such as CLIP and DINO have been successfully adapted for restoration tasks, with DA-CLIP\citep{luo2023controlling} introducing degradation-aware adaptations and Perceive-IR\citep{zhang2025perceive} employing multi-level quality-driven prompts for enhanced semantic understanding. The emergence of agentic restoration systems powered by multimodal large language models marks a paradigm shift toward intelligent, reasoning-based approaches. RestoreAgent\citep{chen2024restoreagent} and AgenticIR\citep{zhu2024intelligent} demonstrate autonomous degradation analysis and adaptive model selection, representing a departure from fixed restoration pipelines toward dynamic systems capable of mimicking human-like problem-solving workflows. These developments collectively indicate a trajectory toward more intelligent, adaptive, and generalizable restoration frameworks that can effectively handle the complexity and unpredictability of real-world image degradations.

\subsection{Remote Sensing Image Restoration}

Remote sensing image restoration has undergone a paradigm shift from task-specific solutions addressing individual degradation types to unified frameworks capable of handling multiple degradation factors simultaneously. Traditional approaches typically focused on single restoration tasks such as denoising, dehazing, deblurring, or illumination enhancement, each requiring specialized architectures and training procedures\citep{he2015total}. However, real-world remote sensing images often suffer from multiple, concurrent degradation factors including atmospheric scattering, sensor noise, motion blur, and varying illumination conditions\citep{vivone2014critical}. This reality has motivated the development of all-in-one restoration frameworks that can adaptively address diverse degradation types within a single, unified architecture, eliminating the need for task-specific models and enabling more efficient processing of operationally acquired remote sensing data\citep{dewangan2024imu}.

Model-driven approaches have emerged as a promising direction for unified remote sensing image restoration, integrating physical degradation models with deep learning architectures to enhance both performance and interpretability. Ada4DIR represents a significant advancement in this domain, proposing an adaptive model-driven all-in-one network that combines four distinct degradation models with a degradation-driven fusion transformer block (4DFTB)\citep{lihe2025ada4dir}. This approach leverages prompt learning mechanisms to guide the restoration process based on degradation type identification, while model-driven restoration modules transform degraded features into clean representations within the feature space. Similarly, progressive restoration networks based on high-order degradation imaging models have demonstrated superior performance by incorporating mathematical interpretability through deep unfolding frameworks\citep{feng2024progressive}. These methods address the limitations of conventional first-order degradation models by progressively handling denoising, deblurring, and super-resolution tasks through theoretically grounded architectures.

Multi-degradation modeling and interaction learning strategies have proven essential for addressing the complex degradation patterns inherent in remote sensing imagery. The saliency-guided interaction learning (SGIL) framework addresses double-degradation scenarios by simultaneously modeling external environmental degradation and internal sensor noise through pseudo pixel supervision-based saliency analysis and task-aware interaction learning modules\citep{wang2024unified}. Multi-stage frameworks incorporating "denoising-deblurring-detail enhancement" pipelines with differentiated intermediate supervision have shown effectiveness in preserving inter-stage information flow while addressing the multi-scale characteristics of remote sensing images\citep{zhang2024method}. Furthermore, unified frameworks for joint super-resolution and registration of multi-angle hyperspectral images demonstrate the potential for addressing multiple restoration tasks through rank minimization approaches that exploit complementary information across different viewing angles\citep{zhou2019integrated}.

Recent advances in transformer architectures and diffusion models have opened new avenues for unified remote sensing image restoration, though significant challenges remain in balancing restoration quality with computational efficiency. Frequency-oriented transformers that explore degradation characteristics in the frequency domain have shown promise for addressing non-uniform haze distributions\citep{zhang2024frequency}, while hybrid approaches combining convolutional neural networks with vision transformers offer improved capability for modeling both local texture details and global contextual information\citep{liu2025remote}. Diffusion-based models for remote sensing image fusion and restoration have demonstrated potential for handling complex degradation patterns, though their computational requirements pose challenges for operational deployment\citep{sui2024denoising}. Despite these advances, the development of truly unified restoration frameworks for remote sensing applications continues to face obstacles including task conflicts during multi-degradation training, the need for accurate degradation type identification, and the requirement for maintaining spectral fidelity across diverse imaging conditions and sensor types.

\section{Method}
\label{sec:System_Model}

\begin{figure*}[tbp]
	\centering
	\includegraphics[width=1.0\linewidth]{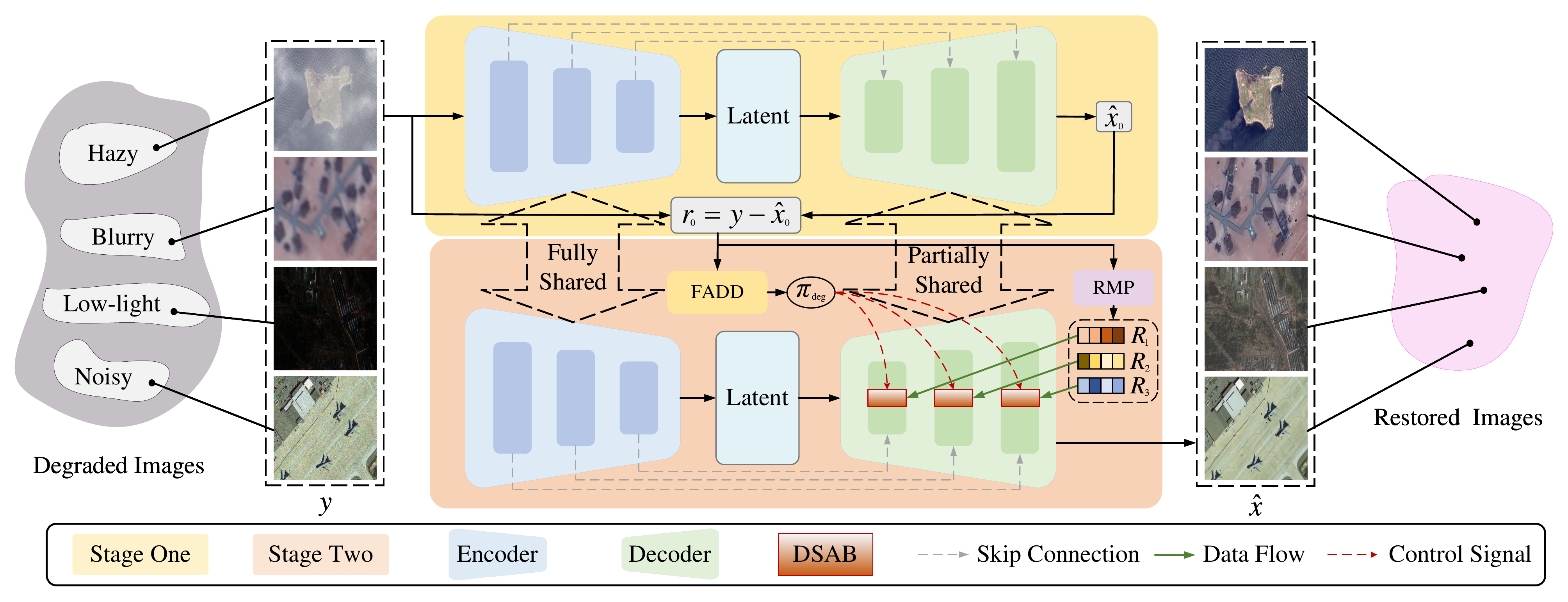}
	\caption{Overall architecture of the proposed PhyDAE framework for all-in-one remote sensing image restoration. The two-stage cascaded pipeline progressively transforms degradation information from implicit features to explicit decision signals through the integration of Frequency-Aware Degradation Decomposer (FADD) and Residual Manifold Projector (RMP) modules.}
	\label{flowchart}
\end{figure*}

\subsection{Overall Framework of PhyDAE}

The fundamental challenge confronting remote sensing image restoration tasks lies in the heterogeneity and physical coupling of degradation processes, where degradation characteristics cannot be adequately modeled and effectively addressed through single degradation assumptions or purely data-driven approaches. To overcome the aforementioned limitations, this paper proposes a Physics-Guided Degradation-Adaptive Experts (PhyDAE) model. As illustrated in Fig.~\ref{flowchart}, the proposed method reformulates the remote sensing image restoration problem as a degradation-aware optimal transport problem with explicit physical constraints.

Let $\mu \in {\cal P}({\cal Y})$ denote the empirical distribution of degraded observations and $\nu \in {\cal P}({\cal X})$ the target distribution of clean images. PhyDAE seeks to learn an adaptive transport mapping ${{\cal T}_\Theta}:{\cal Y} \to {\cal X}$ by minimizing a constrained optimal transport objective:

\begin{multline}
	\mathcal{T}_{\Theta}^*=\arg \min _{\mathcal{T}} \mathbb{E}_{(\mathbf{y}, \mathbf{x})}\Big[
	\mathcal{C}_{\text{DAOT}}(\mathbf{y}, \mathcal{T}(\mathbf{y}), \mathbf{x}) 
	+ \lambda_{\text{phys}} \mathcal{R}_{\text{phys}}(\mathbf{y}, \mathcal{T}(\mathbf{y})) \\
	+ \lambda_{\text{dist}} \mathcal{W}_2\!\left(\mathcal{T}_{\#} \mu, v\right)
	\Big]
\end{multline}where $\Theta$ denotes the learnable network parameters, ${\bf y} \in {\cal Y}$ and ${\bf x} \in {\cal X}$ represent the degraded image and its corresponding clean counterpart, respectively, and $\mathbb{E}{({\bf y},{\bf x})}$ denotes the expectation over training pairs. ${\cal T}({\bf y})$ is the output of the transport mapping applied to the degraded image. $\mathcal{C}_{\text {DAOT }}$ denotes the degradation-aware optimal transport cost function, $\mathcal{R}_{\mathrm{phys}}$ is the physics-consistency regularization term, $\mathcal{W}_2$ represents the 2-Wasserstein distance, and $\mathcal{T}_{\#} \mu$ denotes the push-forward measure of the mapping ${\cal T}$. The coefficients $\lambda_{\text {phys }}, \lambda_{\text {dist }}>0$ are regularization parameters that balance the competing objectives. In contrast to traditional pixel-wise optimization paradigms, this optimal-transport-based mathematical framework revisits the image restoration problem from the perspective of probability measures. By minimizing the geometric distance between distributions, it ensures that the restored results are statistically aligned with the distribution of clean images, while the geometric properties of the Wasserstein distance naturally preserve the spatial structural integrity of the images.

Considering that different degradation types exhibit distinct physical characteristics in the frequency domain, we introduce a physics-consistency regularization term to enforce the spectral properties specific to each degradation:

\begin{align}
	\mathcal{R}_{\mathrm{phys}}(\mathbf{y}, \hat{\mathbf{x}})
	= & \sum_{k=1}^K \pi_k(\mathbf{y}) \cdot \mathcal{D}_k\!\left(\mathbf{r}^{(k)}\right) \notag \\
	= & \sum_{k=1}^K \pi_k(\mathbf{y}) 
	\int_{\Omega_k} |\mathcal{F}[\mathbf{r}](\omega)|^2 w_k(\omega)\, d \omega
\end{align}where $\pi_k(\mathbf{y})$ denotes the posterior probability of a degradation type, $\mathbf{r}=\mathbf{y}-\hat{\mathbf{x}}$ represents the transport residual, $\mathcal{F}[\cdot]$ denotes the Fourier transform, $\Omega_k$ specifies the characteristic frequency band of degradation $k$, and $w_k(\omega)$ is the physics-aware weighting function. The theoretical rationale behind this regularization term lies in the fact that the transport residual $\mathbf{r}$ inherently preserves the essential characteristics of the degradation process in the frequency domain. Since different physical degradation mechanisms exhibit markedly distinct energy distribution patterns in the spectrum, imposing physics-driven constraints on specific frequency bands ensures that the restoration process adheres to the corresponding physical laws and spectral properties from a signal processing perspective.

To address the complexity and training instability inherent in end-to-end mapping, PhyDAE employs a cascaded refinement strategy, decomposing the challenging restoration process into two collaboratively solved subproblems. In the first stage, a standard encoder–decoder network $\mathcal{G}^{(0)}$ is used to generate a preliminary estimate from the degraded input image $\mathbf{y}$:

\begin{equation}
	\hat{\mathbf{x}}_0=\mathcal{G}^{(0)}(\mathbf{y})
\end{equation}

The primary goal of the first stage is to remove the dominant degradation components and recover the basic structural information of the image. However, due to the lack of targeted degradation modeling, its output often exhibits issues such as blurred details and residual degradations. Based on the preliminary restoration result, PhyDAE computes the transmission residual $\mathbf{r}_0=\mathbf{y}-\hat{\mathbf{x}}_0$, which encapsulates rich degradation characteristics and error patterns arising during the restoration process. In the second stage, these residual cues are fully leveraged, and a degradation-aware expert network $\mathcal{G}^{(1)}$ is employed for precise refinement:

\begin{equation}
	\hat{\mathbf{x}}=\mathcal{G}^{(1)}\left(\mathbf{y} \mid \mathcal{M}_{\mathrm{RMP}}\left(\mathbf{r}_0\right), \mathcal{F}_{\mathrm{FADD}}\left(\mathbf{r}_0\right), \boldsymbol{\pi}_{\mathrm{deg}}\right)
\end{equation}where $\mathcal{M}_{\mathrm{RMP}}$ denotes the residual manifold projector, responsible for mapping the residual $\mathbf{r}_0$ onto a structured low-dimensional manifold to enhance semantic representation; $\mathcal{F}_{\mathrm{FADD}}$ represents the frequency-aware degradation decomposer, which extracts degradation-specific spectral features via multi-scale frequency analysis; and $\boldsymbol{\pi}_{\mathrm{deg}}$ corresponds to the degradation posterior distribution obtained from residual analysis. This cascaded design enables PhyDAE, in the second stage, to focus on detail restoration and degradation-specific fine processing, while the residual guidance ensures that the refinement process remains targeted and effective.

\subsection{Residual-Guided Degradation Analysis}

Traditional image restoration methods often overlook the rich degradation information embedded in the transmission residual, which is crucial for accurately understanding and modeling the degradation process. Although the transmission residual $\mathbf{r}_0$ encodes abundant degradation cues, it exhibits complex nonlinear structures in the pixel space, limiting its direct utility. To fully exploit this information, we design two complementary analysis modules: the Residual Manifold Projector (RMP) and the Frequency-Aware Degradation Decomposer (FADD), which perform in-depth analysis of the residual from the perspectives of manifold geometry and frequency characteristics, respectively.

\begin{figure}[tbp]
	\centering
	\includegraphics[width=1.0\linewidth]{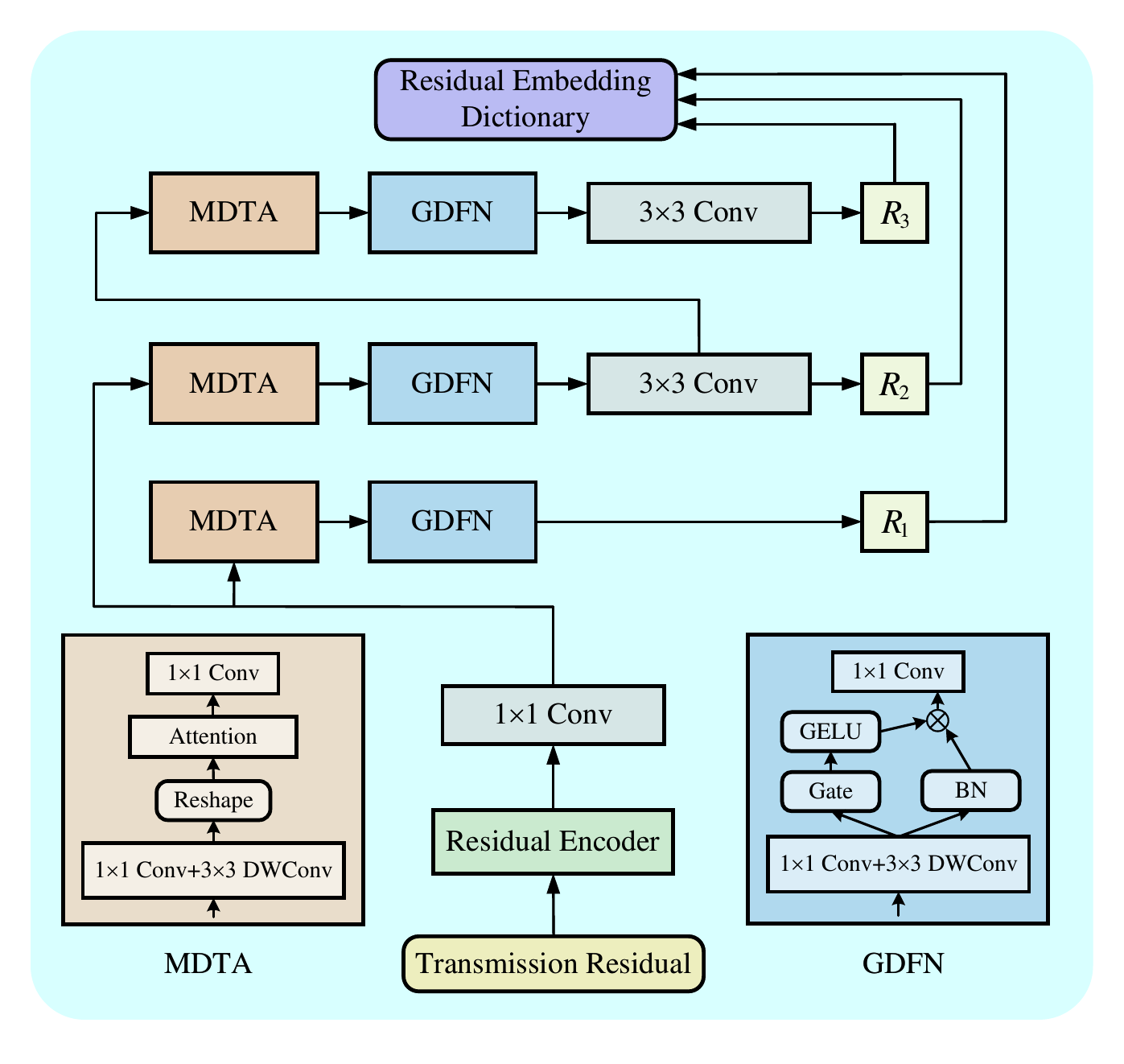}
	\caption{Detailed architecture of the Residual Manifold Projector (RMP) module. The hierarchical structure incorporates Multi-Depthwise Convolution Head Transposed Attention (MDTA) and Gated Depthwise Convolution Feed-Forward Network (GDFN) to extract multi-scale residual embeddings from manifold geometry perspective.}
	\label{RMP}
\end{figure}

According to manifold learning theory, residuals corresponding to different degradation types often reside on distinct manifolds in high-dimensional space. Accordingly, RMP is designed to learn the intrinsic structures of these manifolds, thereby providing a more effective representation of degradation characteristics. As illustrated in Fig.~\ref{RMP}, RMP first employs a residual encoder $\mathcal{E}_{\mathrm{res}}$ to extract the basic features:

\begin{equation}
	\mathbf{R}_0=\mathcal{E}_{\mathrm{res}}\left(\mathbf{r}_0\right)
\end{equation}

Multi-scale residual embeddings are then generated through the designed cascaded attention mechanism:

\begin{equation}
	\begin{aligned}
		& \mathbf{R}_1=\operatorname{GDFN}_1\left(\operatorname{MDTA}_1\left(\operatorname{Conv}_{1 \times 1}\left(\mathbf{R}_0\right)\right)\right) \\
		& \mathbf{R}_2=\operatorname{Conv}_{3 \times 3}\left(\operatorname{GDFN}_2\left(\operatorname{MDTA}_2\left(\operatorname{Conv}_{1 \times 1}\left(\mathbf{R}_0\right)\right)\right)\right) \\
		& \mathbf{R}_3=\operatorname{Conv}_{3 \times 3}\left(\operatorname{GDFN}_3\left(\operatorname{MDTA}_3\left(\operatorname{Conv}_{1 \times 1}\left(\mathbf{R}_2\right)\right)\right)\right)
	\end{aligned}
\end{equation}

Within this hierarchical architecture, the Multi-Depthwise Convolution Head Transposed Attention (MDTA) module achieves computationally efficient spatial attention through depthwise separable convolutions:

\begin{equation}
	\operatorname{MDTA}(\mathbf{X})=\operatorname{GroupNorm}\left(\operatorname{Conv}_{1 \times 1}\left(\operatorname{DWConv}_{3 \times 3}\left(\operatorname{Conv}_{1 \times 1}(\mathbf{X})\right)\right)\right)
\end{equation}

This design maintains the capability to model local spatial relationships while substantially reducing computational complexity through depthwise separable operations. The Gated Depthwise Convolution Feed-Forward Network (GDFN) adopts a similar architecture but incorporates GELU activation functions to provide non-linear transformation capability, enabling selective feature enhancement. To ensure effective utilization of residual information across different decoder levels, the RMP generates dimension-adaptive embedding representations:

\begin{equation}
	\mathbf{E}^{(d)}=\operatorname{Conv}_{1 \times 1}^{(d)}\left(\mathbf{R}_{i \bmod 3}\right)
\end{equation}where $d \in \mathcal{D}_{\text {decoder }}$ denotes the specific dimensional requirement of the decoder.

\begin{figure}[tbp]
	\centering
	\includegraphics[width=1.0\linewidth]{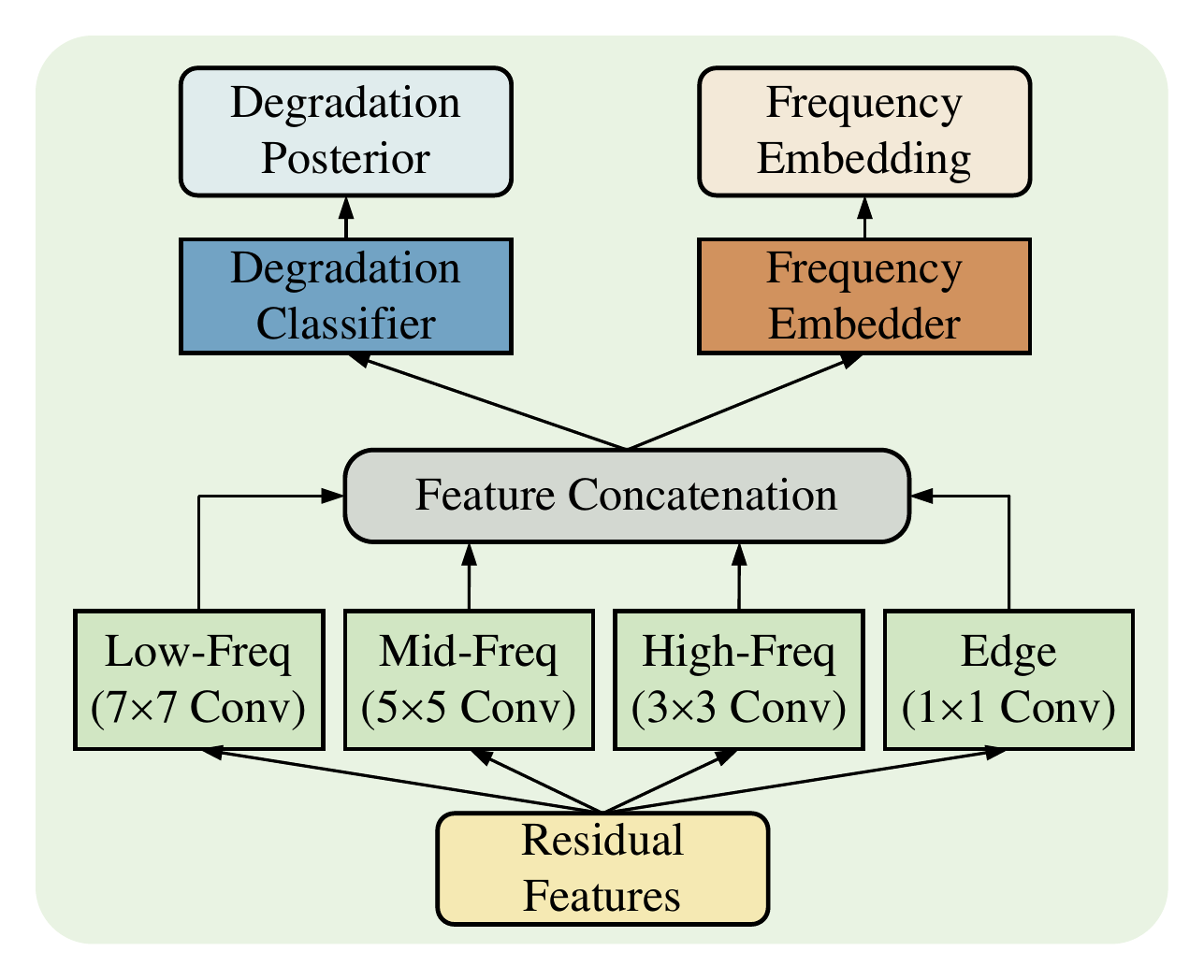}
	\caption{Internal structure of the Frequency-Aware Degradation Decomposer (FADD) module. The multi-scale frequency decomposition strategy employs convolutional kernels with varying receptive fields to capture degradation-specific spectral signatures for posterior probability estimation.}
	\label{FADD}
\end{figure}

Unlike RMP, which analyzes residuals from a spatial structure perspective, FADD focuses on extracting frequency-domain features. Different degradation types in remote sensing imaging exhibit distinct characteristic patterns in the frequency domain: atmospheric scattering primarily attenuates high-frequency details while preserving low-frequency large-scale structures; sensor noise introduces random perturbations across the entire frequency spectrum but manifests more prominently in high-frequency regions; motion blur presents as directional attenuation of frequency responses. Based on these physical properties, as illustrated in Fig.~\ref{FADD}, FADD precisely captures the frequency-domain characteristics of different degradations through a multi-scale frequency decomposition strategy. Specifically, the system employs group convolution kernels with different receptive fields for frequency decomposition: 7$\times$7 kernels extract low-frequency large-scale structural information, 5$\times$5 kernels capture mid-frequency texture features, 3$\times$3 kernels focus on high-frequency details, while 1$\times$1 convolutions extract pixel-wise edge responses. These multi-scale frequency components are integrated through feature concatenation to form a comprehensive frequency-domain representation:

\begin{equation}
	\mathbf{F}_{\text {freq }}=\operatorname{Concat}\left(\left[\mathbf{F}_{\text {low }}, \mathbf{F}_{\text {mid }}, \mathbf{F}_{\text {high }}, \mathbf{F}_{\text {edge }}\right]\right)
\end{equation}

Based on the structured residual representations extracted by RMP, FADD accurately estimates the degradation type distribution of the input image. Through adaptive pooling and fully connected mapping, the degradation posterior probability is generated:

\begin{equation}
	\boldsymbol{\pi}_{\text {deg }}=\operatorname{Softmax}\left(\operatorname{Linear}\left(\operatorname{Flatten}\left(\operatorname{AdaptivePool}\left(\mathbf{R}_1\right)\right)\right)\right)
\end{equation}where $\boldsymbol{\pi}_{\mathrm{deg}} \in \Delta^3$ represents a distribution on the 4-dimensional probability simplex, corresponding to four primary remote sensing image degradation types: haze, noise, low-light, and blur. This probabilistic representation not only handles single degradation scenarios but, more importantly, addresses the mixed degradation problems commonly encountered in practical remote sensing imaging, thereby providing a reliable decision basis for subsequent adaptive expert selection.

\subsection{Physics-Aware Expert Network}

Most all-in-one remote sensing image restoration methods adopt generic neural network architectures to handle various types of degradations. While this "one-network-fits-all" design strategy is relatively simple to implement, it fundamentally overlooks the intrinsic differences among degradation types in terms of physical mechanisms, mathematical modeling, and optimal processing strategies. Based on this insight, PhyDAE employs a mixture-of-experts architecture, constructing specialized processing networks for each major degradation type to ensure physical plausibility and targeted effectiveness in the restoration process.

\begin{figure*}[tbp]
	\centering
	\includegraphics[width=1.0\linewidth]{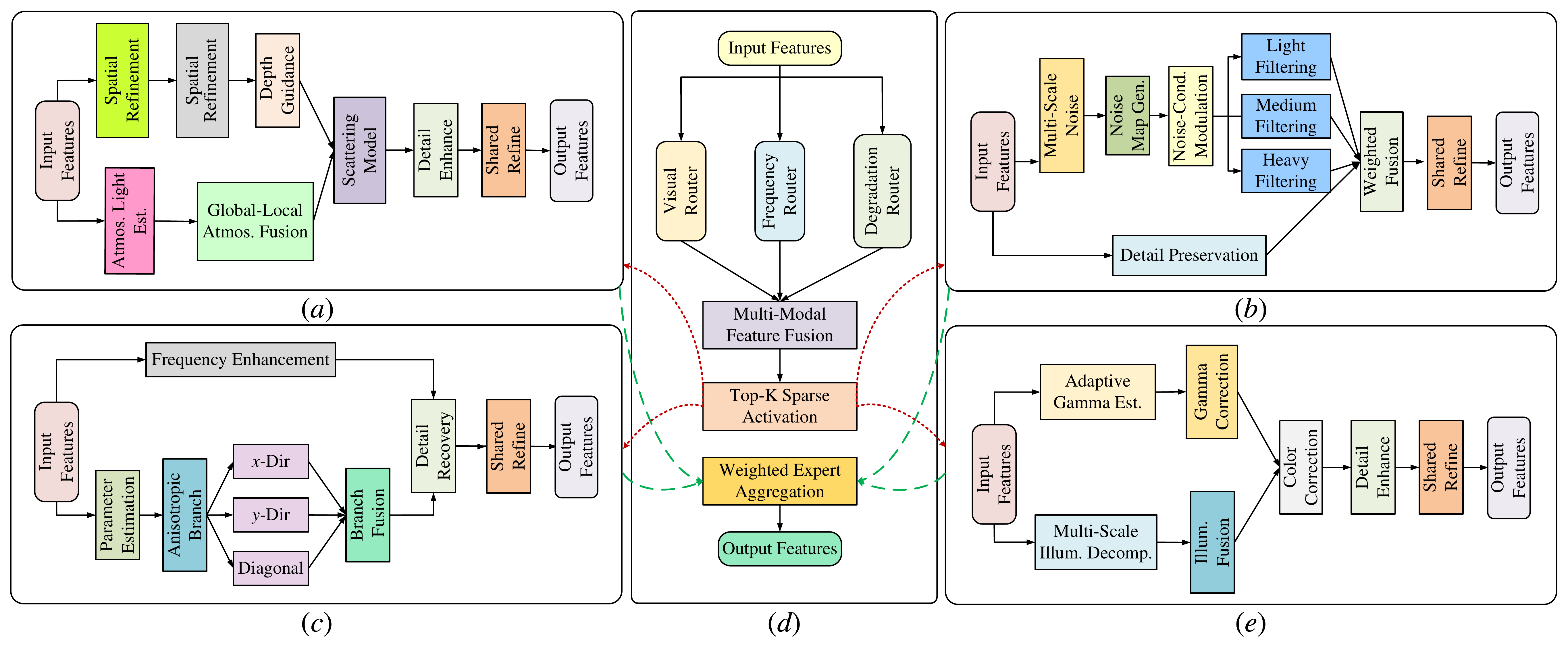}
	\caption{Physics-aware expert networks and temperature-controlled sparse activation mechanism. (a) Dehazing expert incorporating atmospheric scattering model. (b) Denoising expert with spatial-adaptive filtering strategy. (c) Deblurring expert utilizing anisotropic Gaussian modeling. (d) Probabilistic expert allocation with Top-K sparse routing. (e) Low-light enhancement expert based on Retinex theory.}
	\label{expert}
\end{figure*}

The dehazing expert is rigorously designed based on the classical atmospheric scattering physical model. As illustrated in Fig.~\ref{expert}(a), this model characterizes the hazy image formation process as:

\begin{equation}
	\mathbf{I}(\mathbf{p})=\mathbf{J}(\mathbf{p}) \odot \mathbf{t}(\mathbf{p})+\mathbf{A} \odot(1-\mathbf{t}(\mathbf{p}))
\end{equation}where $\mathbf{I}(\mathbf{p})$ denotes the observed hazy image, while $\mathbf{J}(\mathbf{p})$, $\mathbf{t}(\mathbf{p})$, and $\mathbf{A}$ represent the scene radiance to be recovered, the atmospheric transmittance, and the global atmospheric light, respectively. Considering the wavelength-dependent scattering properties of light, the dehazing expert system estimates the transmittance parameter separately for each color channel:

\begin{equation}
	\mathbf{t}_\lambda(\mathbf{p})=\sigma\left(\mathcal{N}_\lambda(\mathbf{X})\right) \cdot s_\lambda
\end{equation}where $\lambda \in\{R, G, B\}$ corresponds to the red, green, and blue color channels, while $S_\lambda$ denotes the wavelength-dependent scaling factor. This processing strategy accurately models the scattering differences of light across different wavelengths in the atmosphere, substantially enhancing the color fidelity of the dehazed results. Upon estimating the transmittance, the expert system employs a spatial refinement mechanism to ensure both spatial consistency and edge sharpness of the transmittance maps. For atmospheric light estimation, the expert network adopts an adaptive strategy that combines global and local information:

\begin{equation}
	{\bf{A}} = {w_{{\mathop{\rm mix}\nolimits} }} \cdot {{\bf{A}}_{{\rm{global }}}} + \left( {1 - {w_{{\mathop{\rm mix}\nolimits} }}} \right) \cdot {{\bf{A}}_{{\rm{local }}}}
\end{equation}where the mixing weight ${w_{{\rm{mix }}}}$ is adaptively adjusted according to the scene complexity and spatial homogeneity. Finally, the scene radiance is precisely recovered through the physical inverse transformation:

\begin{equation}
	\hat{\mathbf{J}}(\mathbf{p})=\frac{\mathbf{I}(\mathbf{p})-\mathbf{A} \odot(1-\mathbf{t}(\mathbf{p}))}{\mathbf{t}(\mathbf{p})+\epsilon}
\end{equation}

The denoising expert, accounting for the complex statistical characteristics of sensor noise, abandons the traditional assumption of globally uniform noise and instead employs a spatially adaptive processing strategy based on locally estimated noise properties. As illustrated in Fig.~\ref{expert}(b), the system first generates high-precision spatial noise maps through a multi-scale noise analysis network. These maps not only reflect the spatial distribution of noise intensity but also distinguish between different noise types. Based on the estimated noise strength, the system applies a three-tier adaptive processing strategy:

\begin{equation}
	{w_s}(\hat \sigma ) = \exp \left( { - \frac{{{{\left( {\hat \sigma  - {\mu _s}} \right)}^2}}}{{2\sigma _s^2}}} \right)
\end{equation}where $s$ corresponds to the denoising schemes for different noise intensity levels. Mild noise processing emphasizes detail preservation and employs a gentle filtering strategy; moderate noise processing seeks a balance between denoising effectiveness and detail retention; while severe noise processing adopts an aggressive denoising strategy to ensure the basic usability of the image. The final output is obtained through a weighted fusion of these results:

\begin{equation}
	\hat{x}=\frac{\sum_s w_s(\hat{\sigma}) \mathcal{F}_s(\mathbf{x})}{\sum_s w_s(\hat{\sigma})+\epsilon}
\end{equation}

The low-light enhancement expert, grounded in Retinex visual theory, models the enhancement of low-light images as a problem of separating and reconstructing illumination and reflectance components. According to this theory, the perceived image can be decomposed into intrinsic reflectance, which is invariant to lighting conditions, and illumination, which varies with the environment. By processing these two components separately, natural illumination enhancement can be achieved. As illustrated in Fig.~\ref{expert}(e), the expert network jointly models local and global illumination features to accurately estimate non-uniform illumination distributions in complex scenes. Specifically, the network first estimates per-pixel local gamma parameters and global illumination intensity, and then applies illumination-aware adaptive gamma correction based on these estimates to improve the overall enhancement effect:
\begin{equation}
	\hat{\mathbf{J}}(\mathbf{p})=\left(\frac{\mathbf{I}(\mathbf{p})}{\mathbf{L}(\mathbf{p})+\epsilon}\right)^{1 / \gamma_{\operatorname{map}}(\mathbf{p})}
\end{equation}where $\mathbf{L}(\mathbf{p})$ denotes the estimated illumination component, and $\gamma_{\operatorname{map}}(\mathbf{p})$ represents the spatially adaptive gamma map. This processing strategy not only effectively enhances the visibility of dark regions but also preserves the natural appearance and color consistency of the image.

The deblurring expert is specifically designed to address image blur caused by platform vibrations and target motion in remote sensing. As illustrated in Fig.~\ref{expert}(c), this expert employs a parameterized anisotropic Gaussian blur model and estimates the blur kernel parameters through a neural network:

\begin{equation}
	\left[\kappa, \sigma_x, \sigma_y, \theta\right]=\mathcal{N}_{\mathrm{param}}(\mathbf{X})
\end{equation}where $\kappa, \sigma_x, \sigma_y, \theta$ corresponds to the blur intensity, horizontal standard deviation, vertical standard deviation, and principal axis orientation, respectively. Based on the anisotropy measure $a=\left|\sigma_x-\sigma_y\right|$, the system introduces directionally adaptive filtering, where the adaptive weights are defined as:

\begin{equation}
	\mathbf{W}_{\text {dir }}=\operatorname{Softmax}\left([1-a, a(1+\cos (\theta \pi)), a(1+\sin (\theta \pi))]^T\right)
\end{equation}

This design embeds the directional information of motion blur into the filter weights, enabling the system to adaptively select the most appropriate restoration filter for blurs along different directions.

To ensure that all experts can fully leverage the results of the aforementioned residual analysis and achieve semantically aligned processing, each expert is conditioned on the residual embeddings provided by the RMP:

\begin{equation}
	\mathbf{x}_{\text {expert }}^{(k)}=\mathcal{E}_k\left(\mathbf{x}, \mathbf{E}^{(d)}\right)
\end{equation}where $\mathbf{E}^{(d)}$ denotes the residual embedding corresponding to the $l$-th layer of the decoder. The significance of this conditional design lies in enabling each expert not only to possess specialized capability for handling a specific degradation type but also to finely adjust its processing based on the degradation characteristics of the current sample, thereby substantially enhancing restoration accuracy and robustness.

\subsection{Probabilistic Expert Allocation and Sparse Activation}

PhyDAE incorporates a comprehensive probabilistic expert allocator, which leverages multiple complementary information sources to achieve more precise and efficient expert selection. Specifically, as illustrated in Fig.~\ref{expert}(d), the system extracts routing features from three complementary perspectives: visual features $\mathbf{h}_{\text {visual }}$, which capture the overall visual information through global adaptive pooling of image features; frequency features $\mathbf{h}_{\mathrm{freq}}$, which encode degradation-specific spectral characteristics derived from the frequency-domain embeddings obtained by the FADD module; and degradation priors $\mathbf{h}_{\mathrm{deg}}$, computed based on the degradation probability distribution obtained from residual analysis:

\begin{equation}
	\begin{aligned}
		\mathbf{h}_{\text {visual }} & =\text { AdaptivePool }\left(\mathcal{N}_{\text {visual }}(\mathbf{X})\right) \\
		\mathbf{h}_{\text {freq }} & =\mathcal{N}_{\text {freq }}\left(\mathbf{f}_{\text {emb }}\right) \\
		\mathbf{h}_{\text {deg }} & =\boldsymbol{\pi}_{\text {deg }}^T \mathbf{W}_{\text {deg }} \\
		\ell & =\mathbf{h}_{\text {visual }}+\mathbf{h}_{\text {freq }}+\alpha \mathbf{h}_{\text {deg }}
	\end{aligned}
\end{equation}where the parameter $\alpha$ controls the influence of prior information on the routing decision, and $\mathbf{W}_{\mathrm{deg}}$ represents the learnable degradation-type embedding matrix. By performing multi-modal fusion, the mixture-of-experts system jointly models sample features from different perspectives, thereby achieving a more comprehensive feature representation and supporting more informed expert selection and routing decisions.

To simultaneously achieve high-precision expert selection and a significant improvement in computational efficiency, PhyDAE employs a temperature-controlled Top-K sparse activation strategy. The core idea of this strategy is to activate only the few most relevant experts during each forward pass, reducing the computational complexity from $\mathcal{O}(K)$, which scales with the total number of experts, to a constant-level cost $\mathcal{O}\left(K_{\text {active }}\right)$:

\begin{equation}
	\begin{aligned}
		\mathbf{w}_{\text {route }} & =\operatorname{Softmax}\left(\frac{\ell}{\tau}\right) \\
		\left(\mathbf{w}_{\text {top }}, \mathbf{I}_{\text {top }}\right) & =\operatorname{TopK}\left(\mathbf{w}_{\text {route }}, K\right) \\
		\mathbf{y}_{\text {expert }} & =\sum_{k=1}^K \frac{w_{\text {top }}^{(k)}}{\sum_{j=1}^K w_{\text {top }}^{(j)}} \mathcal{E}_{I_{\text {top }}^{(k)}}\left(\mathbf{x}, \mathbf{E}^{(d)}\right)
	\end{aligned}
\end{equation}

Within the routing mechanism, the temperature parameter $\tau$ regulates the concentration of expert selection. Lower temperature values concentrate the selection on a few high-weight experts, thereby improving computational efficiency, whereas higher temperatures encourage the participation of more experts, enhancing model expressiveness at the cost of increased computation. By appropriately setting $\tau$, the system can achieve a balance between accuracy and efficiency. Moreover, the number of experts $K$ activated per sample can be flexibly adjusted according to application requirements and computational resources. During training, a larger $K$ is typically employed to fully capture diverse degradation patterns, while during inference, $K$ is reduced to improve efficiency. Additionally, a dynamic routing strategy is introduced, allowing the number of activated experts to adaptively adjust based on sample complexity, thereby balancing computational cost and restoration quality.

\subsection{Multi-Dimensional Joint Optimization Strategy}

The multi-faceted nature of the all-in-one remote sensing image restoration task makes it difficult for a single loss function to comprehensively evaluate restoration quality. To address this, PhyDAE designs a multi-dimensional joint optimization framework, wherein the core Degradation-Aware Optimal Transport (DAOT) loss is formulated as:

\begin{equation}
	\mathcal{L}_{\text {DAOT }}=\mathcal{W}_2\left(\mathcal{T}_{\#} \mu_{\mathbf{y}}, \nu_{\mathbf{x}}\right)+\lambda_{\text {freq }} \sum_{k=1}^K \pi_k \mathcal{R}_{\text {freq }}^{(k)}(\mathbf{r})
\end{equation}

This design incorporates the Wasserstein distance to preserve distributional and geometric consistency, and introduces physical constraints via task-specific frequency-domain weighting, thereby ensuring both modeling fidelity and restoration stability across different degradation types.

Considering the variability of restoration objectives among different degradation types, a task-adaptive pixel-wise loss mechanism is devised:

\begin{equation}
	\mathcal{L}_{\text {pixel }}=\sum_{i=1}^B\left(\beta_1^{\left(k_i\right)} \mathcal{L}_{\mathrm{L} 1}+\beta_2^{\left(k_i\right)} \mathcal{L}_{\mathrm{FFT}}+\beta_3^{\left(k_i\right)} \mathcal{L}_{\mathrm{SSIM}}\right)
\end{equation}where $k_i$ denotes the degradation type of the $i$-th sample, and the task-specific weight $\left\{\beta_j^{(k)}\right\}_{j=1}^3$ is adaptively adjusted via meta-learning to match the sensitivity of different degradations to each loss component, thereby enabling targeted restoration optimization.

While the mixture-of-experts architecture enhances representational capacity, it may lead to imbalanced expert utilization: a few experts become over-activated while others remain idle, resulting not only in wasted computational resources but also in the risk of the model degenerating into a single-expert system. To mitigate this issue, a coefficient-of-variation-based expert load balancing loss is introduced:

\begin{equation}
	\mathcal{L}_{\text {balance }}=\frac{\sigma(\overline{\mathbf{w}})^2}{\mu(\overline{\mathbf{w}})^2+\epsilon}, \quad \overline{\mathbf{w}} \in \mathbb{R}^K
\end{equation}where $\overline{\mathbf{w}}$ denotes the average activation rate of each expert within a batch. This loss promotes a uniform distribution of expert usage by minimizing the coefficient of variation, thereby enhancing PhyDAE’s generalization capability and improving restoration performance on rare degradation types.

Furthermore, to enhance the model’s discriminative capability for different degradation types, PhyDAE employs supervised contrastive learning in the residual embedding space:

\begin{equation}
	\mathcal{L}_{\text {contrast }}=-\frac{1}{B} \sum_{i=1}^B \log \frac{\sum_{j:y_j=y_i, j \neq i} \exp \left(\mathbf{z}_i^T \mathbf{z}_j / \tau_c\right)}{\sum_{j \neq i} \exp \left(\mathbf{z}_i^T \mathbf{z}_j / \tau_c\right)}
\end{equation}where $\mathbf{Z}_i$ denotes the residual feature vector normalized by $\ell_2$, and $\tau_c$ is the temperature parameter. This loss encourages representations of the same degradation type to cluster in the high-dimensional space, while maintaining separation between different degradation types, thereby enhancing degradation recognition capability.

Integrating the aforementioned loss components, the overall optimization objective is realized through a weighted linear combination:

\begin{equation}
	\mathcal{L}_{\text {total }}=\mathcal{L}_{\text {DAOT }}+\lambda_1 \mathcal{L}_{\text {pixel }}+\lambda_2 \mathcal{L}_{\text {balance }}+\lambda_3 \mathcal{L}_{\text {contrast }}
\end{equation}where $\lambda_1$, $\lambda_2$, and $\lambda_3$ are set to 1.0, 0.01, and 0.1, respectively.

\section{Experimental Results and Analysis}

\subsection{Datasets}

To comprehensively evaluate the performance of our proposed unified remote sensing image restoration algorithm across various degradation types and application scenarios, we constructed two multi-degradation datasets, MD-RSID and MD-RRSHID. We also employed the MDRS-Landsat\citep{lihe2025ada4dir} benchmark for supplementary assessment. All three datasets address four typical restoration tasks: dehazing, denoising, deblurring, and low-light enhancement. MD-RSID and MD-RRSHID are built from different real-world remote sensing sources, offering diverse scenes and degradation characteristics. The MDRS-Landsat dataset, with its unique physical modeling and band-correlated synthesis strategy, provides an additional dimension for validating the algorithm's generalization capabilities.

(1) \textbf{MD-RSID} dataset is an extension of the RSID\citep{chi2023trinity} remote sensing dehazing benchmark. RSID is the first real-world benchmark for remote sensing image dehazing in military contexts, comprising 1,000 pairs of hazy and corresponding ground-truth clear images. Its hazy images are generated using a physical imaging model that embeds atmospheric scattering theory into a deep learning framework, enabling the joint learning of haze parameters and the synthesis of physically-aware images. Using the high-quality clear images from RSID, we systematically synthesized three other types of degradation. Noise degradatio was simulated by adding zero-mean Gaussian noise to emulate sensor quantization errors and data transmission interference. Blur degradation, which simulates effects from atmospheric turbulence, platform vibration, and relative motion, was generated using randomly varying kernel sizes, diffusion parameters, and rotation angles to ensure diverse blur orientations. Low-light degradation was created via a power function transformation to simulate conditions like nighttime imaging and cloud occlusion. MD-RSID contains 4,000 degraded-clear image pairs, partitioned into training (3,200), validation (400), and test (400) sets at an 8:1:1 ratio.

(2) \textbf{MD-RRSHID} dataset is a multi-degradation extension of the RRSHID\citep{zhu2025real} real-world hazy remote sensing image dataset. As the first large-scale benchmark of its kind, RRSHID contains 3,053 hazy-clear image pairs obtained in collaboration with meteorological agencies, covering diverse atmospheric conditions across various geographical regions in China. The dataset exhibits significant diversity: its geographical coverage includes urban, agricultural, mountainous, and aquatic environments; atmospheric conditions range from light to heavy haze; and its spectral properties display complex, real-world color distortions and spectral shifts. Following the same synthesis principles as MD-RSID, we applied systematic, multi-type degradations to the clear images from RRSHID. The degradation parameters were carefully chosen to reflect the distribution of artifacts observed under practical imaging conditions. The resulting MD-RRSHID dataset consists of 3,053 image pairs, split into training (2,442), validation (305), and test (306) sets at an 8:1:1 ratio.

(3)\textbf{MDRS-Landsat} dataset is a multi-degradation benchmark dataset for remote sensing images created by LiHe et al.\citep{lihe2025ada4dir}. It was built using 5,500 high-quality images ($512 \times 512$ pixels) from the RSHaze\citep{song2023vision} dataset, upon which four types of degradation were systematically synthesized: blur, noise, haze, and low-light. A unique aspect of this dataset is its consideration of the physical properties of remote sensing imaging. Notably, for haze synthesis, it utilizes multi-band information from the Landsat 8 satellite to generate band-correlated, non-uniform haze. MDRS-Landsat includes 5,130 training, 100 validation, and 270 test image pairs, offering a physics-based evaluation benchmark for multi-degradation restoration algorithms.

\subsection{Evaluation Metrics}

To objectively evaluate the image restoration performance of our proposed PhyDAE, we employ three metrics: Peak Signal-to-Noise Ratio (PSNR), Structural Similarity Index Measure (SSIM), and Learned Perceptual Image Patch Similarity (LPIPS). These metrics provide a comprehensive quantification of the algorithm's performance in terms of pixel accuracy, structural fidelity, and perceptual quality, respectively.

PSNR is a widely used objective metric that assesses algorithm performance by measuring the pixel-level reconstruction accuracy between the restored and ground-truth images. It is based on the Mean Squared Error (MSE) and is mathematically defined as:
\begin{equation}
	\text{PSNR} = 10 \cdot \log_{10} \left( \frac{\text{MAX}_I^2}{\text{MSE}} \right)
\end{equation}where $\text{MAX}_I$ represents the maximum possible pixel value of the image, and MSE is the mean squared error between the restored and ground-truth images. A higher PSNR value indicates a smaller pixel-level reconstruction error and thus less image distortion.

SSIM is designed based on properties of the human visual system (HVS). It quantifies image quality by assessing similarities in luminance, contrast, and structure, thereby aligning better with subjective visual perception than PSNR. It is formulated as:
\begin{equation}
	\text{SSIM}(x,y) = \frac{(2\mu_x\mu_y + C_1)(2\sigma_{xy} + C_2)}{(\mu_x^2 + \mu_y^2 + C_1)(\sigma_x^2 + \sigma_y^2 + C_2)}
\end{equation}In this equation, $\mu_x$ and $\mu_y$ denote the local means of images $x$ and $y$, while $\sigma_x$ and $\sigma_y$ are their respective standard deviations. $\sigma_{xy}$ is the covariance, and $C_1$ and $C_2$ are constants for numerical stability. SSIM values range from 0 to 1, where a value closer to 1 signifies higher structural similarity.

LPIPS measures the perceptual difference between images by leveraging hierarchical features from a deep convolutional neural network, which more accurately reflects human visual perception. The metric computes a weighted distance in the feature space by extracting features from multiple layers of a pre-trained network:
\begin{equation}
	\operatorname{LPIPS}(x, y)=\sum_l \frac{1}{H_l W_l}\left\|w_l \odot\left(\hat{f}_l(x)-\hat{f}_l(y)\right)\right\|_2^2
\end{equation}where $\hat{f}_l(\cdot)$ denotes the normalized feature map of the $l$-th layer, $H_l$ and $W_l$ represent the spatial dimensions of the feature map, and $w_l$ is a learnable channel-wise weighting coefficient. A lower LPIPS value indicates higher perceptual similarity.

\subsection{Implementation Details}

All experiments were conducted on a distributed cluster equipped with eight NVIDIA A800 GPUs, each with 80GB of VRAM. Our model is implemented using the PyTorch Lightning framework and trained across multiple GPUs via the Distributed Data Parallel (DDP) strategy. We trained the model for 200 epochs using the AdamW optimizer. The initial learning rate was set to $1 \times 10^{-4}$ with a weight decay of $1 \times 10^{-2}$. A learning rate scheduler was employed, combining a linear warmup for the first 10 epochs with a subsequent cosine annealing schedule down to a minimum of $1 \times 10^{-6}$. We used a per-GPU batch size of 8, resulting in a global batch size of 64. All input images were resized to a resolution of $256 \times 256$ pixels. To enhance model generalization, we applied data augmentation techniques during training, including random horizontal flipping and random rotations. For validation, we used only center cropping to ensure consistent evaluation.

The initial embedding dimension of the PhyDAE network is set to $d=32$. The architecture follows a hierarchical design where each downsampling stage doubles the feature dimensions while halving the spatial resolution. The encoder is a 4-level architecture with [4, 6, 6, 8] residual groups per level. The decoder contains [2, 4, 4] dual-stream attention modules. The number of attention heads in the multi-head self-attention mechanism is set to [1, 2, 4, 8] for the different resolution levels. The Sparse Adaptive Expert Adapter utilizes a low-rank decomposition with $r=8$ and employs 4 experts, each specialized for a task: dehazing, denoising, low-light enhancement, and deblurring. The Top-K routing is configured with k=1 for computational efficiency. The base dimension of the Residual Manifold Projector is aligned with the encoder, while its target dimensions match those of the respective decoder layers.

\begin{table*}[tbp]
	\centering
	\caption{Quantitative comparison of all-in-one image restoration methods on MD-RSID dataset across four degradation types.}
	\label{tab:RSID}
	\setlength{\tabcolsep}{4pt}
	\renewcommand{\arraystretch}{1.2}
	\resizebox{\textwidth}{!}{
		\begin{tabular}{@{}lcccccccccccc@{}}
			\toprule
			\multirow{2.5}{*}{\textbf{Method}} & \multicolumn{3}{c}{\textbf{Dehazing}} & \multicolumn{3}{c}{\textbf{Deblurring}} & \multicolumn{3}{c}{\textbf{Denoising}} & \multicolumn{3}{c}{\textbf{Low-light Enhancement}} \\
			\cmidrule(lr){2-4} \cmidrule(lr){5-7} \cmidrule(lr){8-10} \cmidrule(lr){11-13}
			& PSNR↑ & SSIM↑ & LPIPS↓ & PSNR↑ & SSIM↑ & LPIPS↓ & PSNR↑ & SSIM↑ & LPIPS↓ & PSNR↑ & SSIM↑ & LPIPS↓ \\
			\midrule
			AirNet      & 19.79 & 0.8900 & 0.1420 & 21.40 & 0.6526 & 0.4380 & 31.28 & 0.8714 & 0.2370 & 24.25 & 0.9183 & 0.0755 \\
			PromptIR    & 20.00 & 0.8992 & 0.1212 & 22.39 & 0.6510 & 0.4374 & \underline{32.71} & \underline{0.8858} & \underline{0.2256} & 24.04 & 0.9175 & 0.0704 \\
			MPMF-Net    & 19.12 & 0.8868 & 0.1587 & 20.97 & 0.6002 & 0.4588 & 28.56 & 0.8308 & 0.2895 & 24.35 & 0.9082 & 0.0609 \\
			MWFormer    & 22.81 & 0.8657 & 0.1603 & 25.44 & 0.6763 & 0.4264 & 29.87 & 0.7929 & 0.2955 & 28.54 & 0.9528 & 0.0682 \\
			AdaIR       & 22.21 & 0.8989 & 0.1431 & 26.02 & 0.7309 & 0.3861 & 30.76 & 0.8373 & 0.2719 & 28.01 & 0.9572 & 0.0553 \\
			Ada4DIR     & 21.95 & 0.8992 & 0.1277 & 24.19 & 0.7099 & 0.3889 & 28.78 & 0.7796 & 0.2957 & 24.61 & 0.9326 & 0.0695 \\
			OneRestore  & \underline{25.54} & 0.9061 & 0.1249 & 25.89 & 0.6896 & 0.4162 & 30.48 & 0.8195 & 0.2802 & \underline{29.17} & \underline{0.9699} & 0.0441 \\
			GridFormer  & 20.12 & 0.8762 & 0.1296 & 26.13 & 0.7185 & \underline{0.3300} & 30.39 & 0.8221 & 0.2493 & 27.22 & 0.9593 & \underline{0.0284} \\
			DA-RCOT     & 23.29 & 0.9219 & 0.1185 & \underline{26.33} & \underline{0.7369} & 0.3771 & 31.08 & 0.8436 & 0.2696 & 27.56 & 0.9642 & 0.0469 \\
			AutoDIR     & 24.53 & \underline{0.9386} & \underline{0.0983} & 26.28 & 0.7233 & 0.3901 & 29.49 & 0.7687 & 0.3002 & 27.24 & 0.9615 & 0.0499 \\
			\midrule
			\textbf{PhyDAE} & \textbf{26.86} & \textbf{0.9613} & \textbf{0.0586} & \textbf{27.73} & \textbf{0.7772} & \textbf{0.3221} & \textbf{32.77} & \textbf{0.8862} & \textbf{0.2136} & \textbf{31.96} & \textbf{0.9855} & \textbf{0.0211} \\
			\bottomrule
		\end{tabular}
	}
	
	\vspace{0mm}
	\footnotesize
	\noindent * Best results are in \textbf{bold}, second-best results are \underline{underlined}. ↑/↓ denote higher/lower is better.
\end{table*}

\begin{table*}[tbp]
	\centering
	\caption{Quantitative comparison of all-in-one image restoration methods on MD-RRSHID dataset across four degradation types.}
	\label{tab:RRSHID}
	\setlength{\tabcolsep}{4pt}
	\renewcommand{\arraystretch}{1.2}
	\resizebox{\textwidth}{!}{
		\begin{tabular}{@{}lcccccccccccc@{}}
			\toprule
			\multirow{2.5}{*}{\textbf{Method}} & \multicolumn{3}{c}{\textbf{Dehazing}} & \multicolumn{3}{c}{\textbf{Deblurring}} & \multicolumn{3}{c}{\textbf{Denoising}} & \multicolumn{3}{c}{\textbf{Low-light Enhancement}} \\
			\cmidrule(lr){2-4} \cmidrule(lr){5-7} \cmidrule(lr){8-10} \cmidrule(lr){11-13}
			& PSNR↑ & SSIM↑ & LPIPS↓ & PSNR↑ & SSIM↑ & LPIPS↓ & PSNR↑ & SSIM↑ & LPIPS↓ & PSNR↑ & SSIM↑ & LPIPS↓ \\
			\midrule
			AirNet & 20.74 & 0.5187 & 0.5809 & 22.69 & 0.6606 & 0.3688 & 33.83 & 0.8834 & 0.2891 & 27.01 & 0.8422 & 0.3020 \\
			PromptIR & 19.72 & 0.5049 & 0.5585 & 23.83 & 0.7049 & 0.3640 & \textbf{35.46} & \textbf{0.9182} & 0.2607 & \underline{36.52} & 0.9699 & 0.0842 \\
			MPMF-Net & 18.70 & 0.5168 & 0.5966 & 27.50 & 0.7552 & 0.2999 & 25.60 & 0.8163 & 0.3105 & 34.62 & 0.9247 & 0.0851 \\
			MoCE-IR & \underline{22.86} & \underline{0.6479} & 0.4685 & 33.12 & \underline{0.8795} & 0.2502 & 35.04 & 0.9089 & \underline{0.2396} & 35.31 & \underline{0.9797} & 0.0591 \\
			MWFormer & 18.63 & 0.5093 & 0.5920 & 25.29 & 0.7113 & 0.4527 & 25.80 & 0.5345 & 0.4926 & 28.16 & 0.8644 & 0.1733 \\
			AdaIR & 22.58 & 0.6369 & 0.4690 & 32.51 & 0.8676 & 0.2599 & 34.80 & 0.9016 & 0.2485 & 34.24 & 0.9729 & 0.0636 \\
			Ada4DIR & 19.97 & 0.5148 & 0.5558 & 27.76 & 0.7679 & 0.3071 & 25.22 & 0.5361 & 0.4258 & 24.19 & 0.8200 & 0.2784 \\
			OneRestore & 20.04 & 0.5870 & 0.5744 & 23.62 & 0.7219 & 0.3693 & 31.40 & 0.8251 & 0.3015 & 28.26 & 0.6435 & 0.2049 \\
			GridFormer & 21.46 & 0.5757 & 0.4584 & 31.94 & 0.8465 & \underline{0.2248} & 33.28 & 0.8509 & 0.2543 & 33.69 & 0.9665 & \underline{0.0483} \\
			DA-RCOT & 22.80 & 0.6439 & \underline{0.4569} & \textbf{33.86} & \textbf{0.8938} & 0.2275 & 33.09 & 0.8970 & 0.2845 & 35.91 & 0.9785 & 0.0612 \\
			
			\midrule
			PhyDAE & \textbf{22.96} & \textbf{0.6511} & \textbf{0.4423} & \underline{33.73} & 0.8787 & \textbf{0.2122} & \underline{35.17} & \underline{0.9141} & \textbf{0.2289} & \textbf{37.35} & \textbf{0.9852} & \textbf{0.0442} \\
			\bottomrule
		\end{tabular}
	}
	
	\vspace{0mm}
	\footnotesize
	\noindent * Best results are in \textbf{bold}, second-best results are \underline{underlined}. ↑/↓ denote higher/lower is better.
\end{table*}

\subsection{Comparative Experimental Results}

To validate the effectiveness of the proposed PhyDAE model, we conducted a fair comparison against over state-of-the-art all-in-one image restoration methods. This section analyzes the performance of each method across three distinct datasets.

\subsubsection{Analysis of Results on the MD-RSID Dataset}

\begin{figure*}[htbp]
	\centering
	\includegraphics[width=1.0\linewidth]{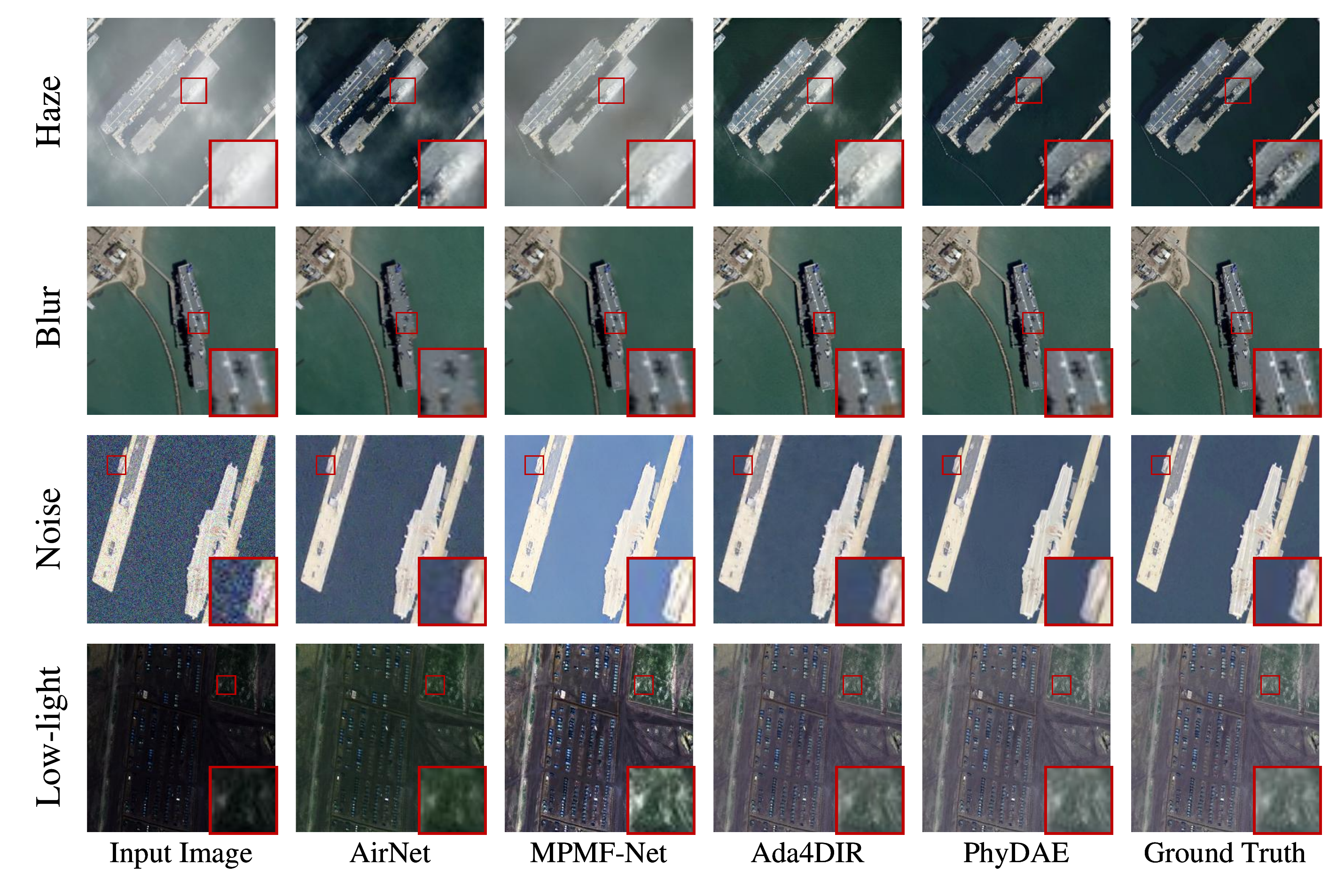}
	\caption{Visual comparison of different all-in-one restoration methods on the MD-RSID dataset. The zoomed-in results are provided to highlight the restoration performance advantages of PhyDAE over state-of-the-art methods.}
	\label{RSID}
	
	\vspace{1em} % 两图之间的间距
	
	\includegraphics[width=1.0\linewidth]{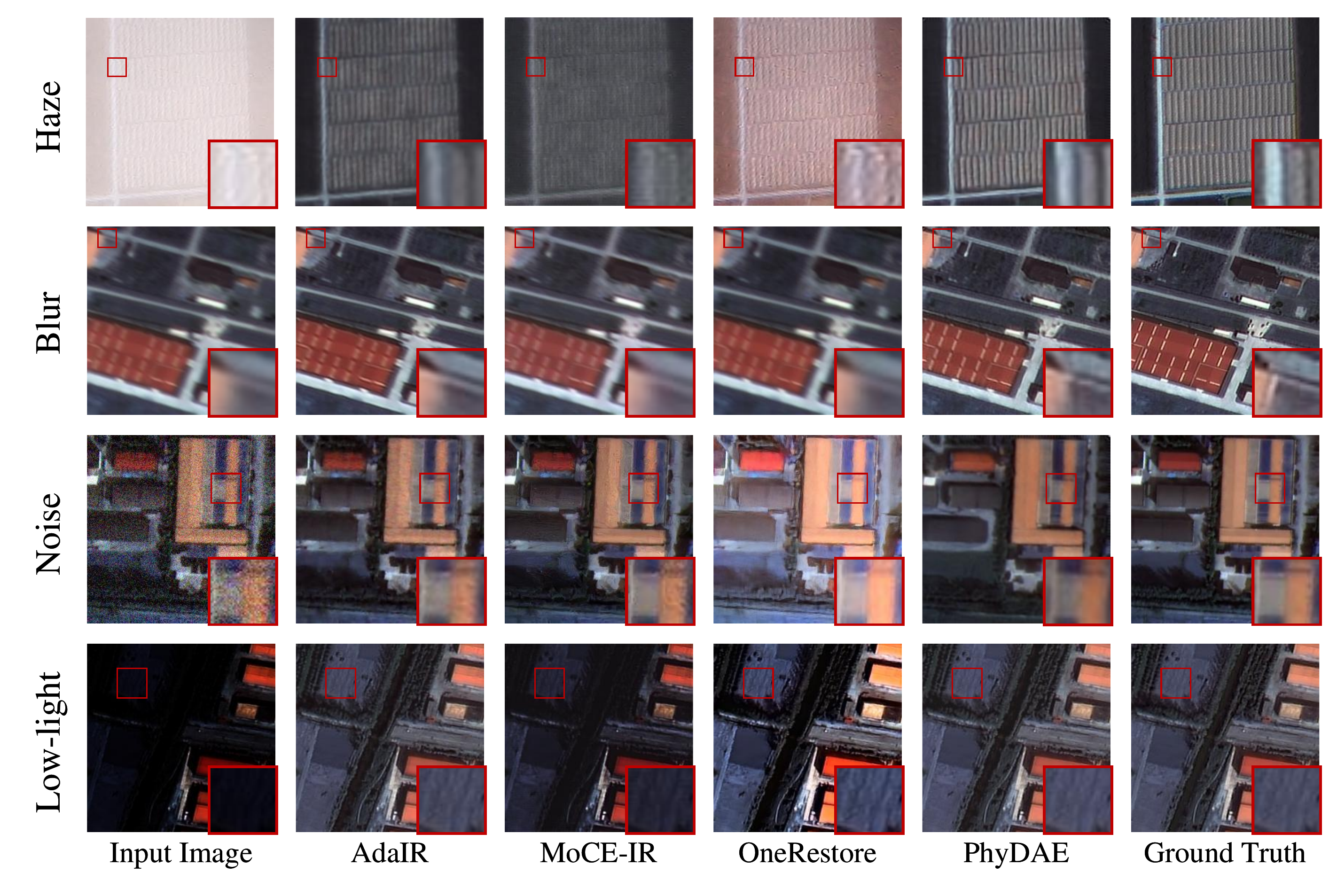}
	\caption{Visual comparison of different all-in-one restoration methods on the MD-RRSHID dataset. The zoomed-in results are provided to highlight the restoration performance advantages of PhyDAE over state-of-the-art methods.}
	\label{RRSHID}
\end{figure*}

\begin{table*}[tbp]
	\centering
	\caption{Quantitative comparison of all-in-one image restoration methods on MDRS-Landsat dataset across four degradation types.}
	\label{tab:MDRS}
	\setlength{\tabcolsep}{4pt}
	\renewcommand{\arraystretch}{1.2}
	\resizebox{\textwidth}{!}{
		\begin{tabular}{@{}lcccccccccccc@{}}
			\toprule
			\multirow{2.5}{*}{\textbf{Method}} & \multicolumn{3}{c}{\textbf{Dehazing}} & \multicolumn{3}{c}{\textbf{Deblurring}} & \multicolumn{3}{c}{\textbf{Denoising}} & \multicolumn{3}{c}{\textbf{Low-light Enhancement}} \\
			\cmidrule(lr){2-4} \cmidrule(lr){5-7} \cmidrule(lr){8-10} \cmidrule(lr){11-13}
			& PSNR↑ & SSIM↑ & LPIPS↓ & PSNR↑ & SSIM↑ & LPIPS↓ & PSNR↑ & SSIM↑ & LPIPS↓ & PSNR↑ & SSIM↑ & LPIPS↓ \\
			\midrule
			NAFNet & 31.56 & 0.9642 & 0.0417 & 33.10 & 0.8120 & 0.3194 & 33.08 & 0.8263 & 0.1872 & 30.40 & 0.9516 & 0.0542 \\
			Restormer & 36.18 & 0.9867 & 0.0124 & 35.23 & 0.8559 & 0.2099 & 34.53 & 0.8589 & 0.1356 & 37.86 & 0.9872 & 0.0110 \\
			DGUNet & 27.45 & 0.9338 & 0.0713 & 29.64 & 0.7822 & 0.3405 & 30.31 & 0.7314 & 0.2491 & 27.15 & 0.9010 & 0.1339 \\
			TransWeather & 35.02 & 0.9689 & 0.0241 & 33.45 & 0.8159 & 0.2868 & 33.69 & 0.8428 & 0.1530 & 36.33 & 0.9705 & 0.0193 \\
			AirNet & 24.39 & 0.9331 & 0.0641 & 28.27 & 0.7887 & 0.3244 & 30.30 & 0.7446 & 0.1918 & 28.38 & 0.9472 & 0.0569 \\
			PromptIR & 37.61 & 0.9897 & \uline{0.0084} & 36.41 & 0.8861 & 0.1557 & \textbf{34.99} & \textbf{0.8729} & \uline{0.1029} & 39.09 & 0.9900 & 0.0084 \\
			IDR & 36.99 & 0.9892 & 0.0087 & \uline{36.57} & \textbf{0.8902} & \uline{0.1498} & 34.88 & 0.8681 & 0.1091 & 35.19 & 0.9865 & 0.0096 \\
			SrResNet-AP & 34.78 & 0.9825 & 0.0186 & 34.63 & 0.8479 & 0.2273 & 34.70 & 0.8620 & 0.1375 & 33.87 & 0.9823 & 0.0165 \\
			Restormer-AP & 37.36 & 0.9888 & 0.0098 & 35.75 & 0.8732 & 0.1800 & 34.96 & \uline{0.8697} & 0.1239 & 37.27 & 0.9885 & 0.0102 \\
			Uformer-AP & 36.06 & 0.9877 & 0.0107 & 34.64 & 0.8488 & 0.2342 & 34.39 & 0.8533 & 0.1334 & 36.58 & 0.9899 & \uline{0.0069} \\
			Ada4DIR-b & \uline{38.79} & \uline{0.9911} & \textbf{0.0065} & 36.40 & \uline{0.8874} & 0.1567 & \uline{34.97} & 0.8688 & 0.1241 & \uline{40.69} & \uline{0.9923} & \textbf{0.0050} \\
			
			\midrule
			PhyDAE & \textbf{39.12} & \textbf{0.9928} & 0.0227 & \textbf{36.88} & 0.8824 & \textbf{0.1487} & 34.53 & 0.8651 & \textbf{0.0991} & \textbf{42.24} & \textbf{0.9949} & 0.0121 \\
			\bottomrule
		\end{tabular}
	}
	
	\vspace{0mm}
	\footnotesize
	\noindent * Best results are in \textbf{bold}, second-best results are \underline{underlined}. ↑/↓ denote higher/lower is better.
\end{table*}

\begin{figure*}[tbp]
	\centering
	
	\includegraphics[width=1.0\linewidth]{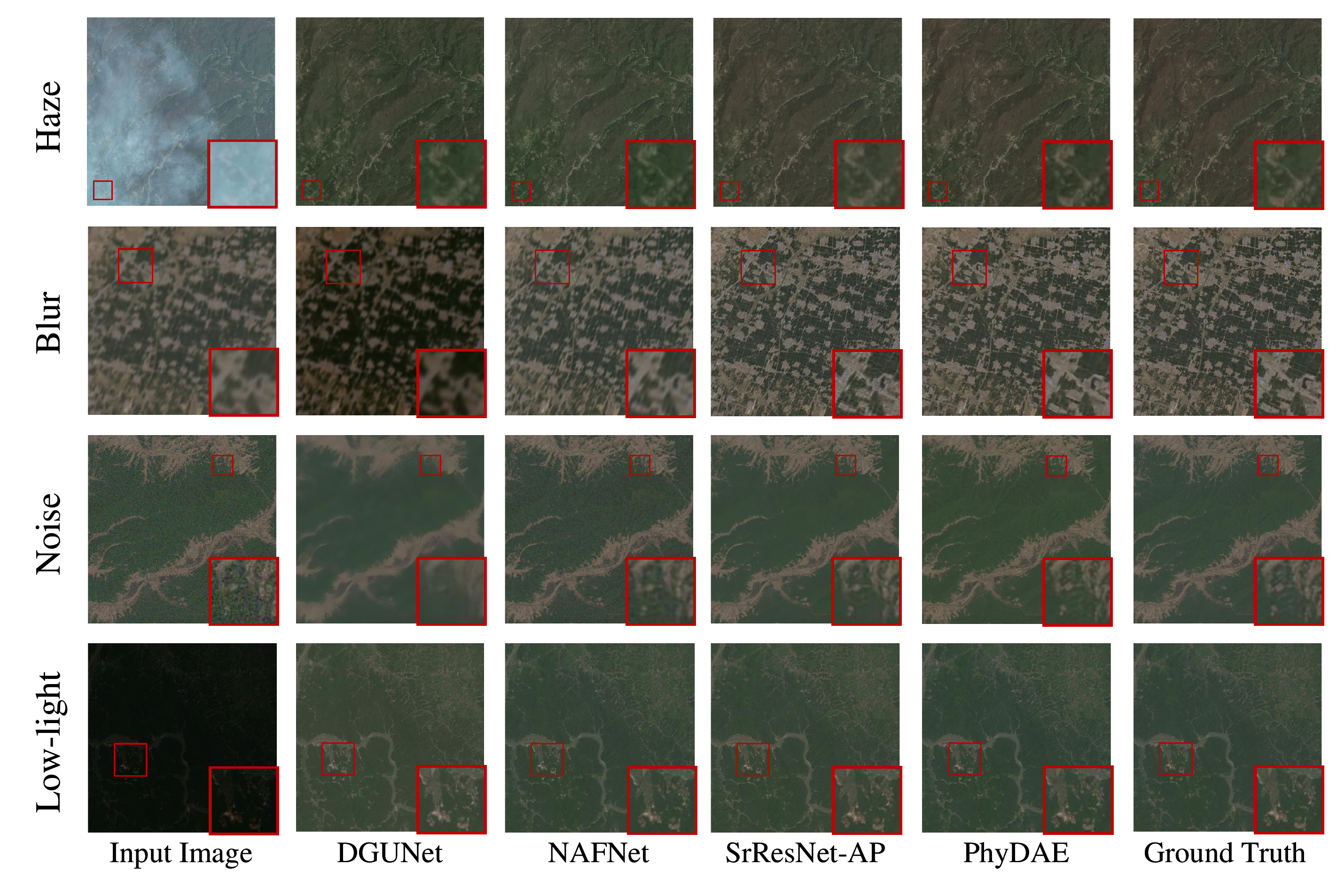}
	\caption{Visual comparison of different all-in-one restoration methods on the MDRS-Landsat dataset. The zoomed-in results are provided to highlight the restoration performance advantages of PhyDAE over state-of-the-art methods.}
	\label{MDRS}
\end{figure*}

Table~\ref{tab:RSID} presents the comparative results on the MD-RSID dataset, where our PhyDAE model demonstrates a significant performance advantage across all four degradation types. Specifically, for dehazing, PhyDAE achieved a PSNR of 26.86 dB and an SSIM of 0.9613, outperforming the second-best method, OneRestore, by 1.32 dB and 0.0552, respectively. Furthermore, it attained the lowest LPIPS score of 0.0586, indicating superior perceptual quality. For deblurring, PhyDAE surpassed all competing methods with a PSNR of 27.73 dB and an SSIM of 0.7772, showcasing robust restoration capabilities, particularly in handling complex motion blur. In the denoising experiments, PhyDAE achieved a leading PSNR of 32.77 dB, exceeding RromptIR (32.71 dB). It also obtained the best SSIM (0.8862) and LPIPS (0.2136) scores, highlighting its comprehensive strengths in both noise suppression and structural information preservation. For low-light enhancement, PhyDAE significantly outperformed other methods with a PSNR of 31.96 dB, which is 2.79 dB higher than the runner-up, OneRestore. Its remarkably low LPIPS score of 0.0211 further confirms its exceptional ability in brightness recovery and detail preservation. The visual comparisons in Fig.~\ref{RSID} show that PhyDAE produces more natural results that are closer to the ground truth, particularly in restoring haze boundaries, motion blur trajectories, noise textures, and details in dark regions.

\subsubsection{Analysis of Results on the MD-RRSHID Dataset}

The MD-RRSHID dataset, constructed from real-world remote sensing imagery, features more complex and diverse degradation patterns. As shown in Table~\ref{tab:RRSHID}, PhyDAE maintained its stable and leading performance on this challenging dataset. For dehazing, it ranked first with a PSNR of 22.96 dB and an SSIM of 0.6511, while its LPIPS score (0.4423) was markedly superior to others, demonstrating its effectiveness in addressing the non-uniform distribution of real-world haze. For deblurring, PhyDAE achieved a PSNR of 33.73 dB. Although this is marginally lower than DA-RCOT's 33.86 dB, it secured the best LPIPS score (0.2122), reflecting the well-balanced performance of the physics-guided mechanism in handling authentic blur. The denoising experiments showcased PhyDAE's exceptional capabilities; its PSNR of 35.17 dB was second only to RromptIR, while its LPIPS of 0.2289 was the best among all competing methods. In low-light enhancement, PhyDAE led by a substantial margin with a PSNR of 37.35 dB (0.83 dB higher than RromptIR) and achieved the top scores for both SSIM (0.9852) and LPIPS (0.0442). Visual results in Fig.~\ref{RRSHID} reveal that PhyDAE better preserves spatial continuity and spectral consistency when processing complex real-world degradations, effectively mitigating over-smoothing and artifacts.

\subsubsection{Analysis of Results on the MDRS-Landsat Dataset}

As an independent, external test set, the MDRS-Landsat dataset provides a crucial validation of the models' generalization capabilities. The results in Table~\ref{tab:MDRS} highlight the performance disparities among the different methods. In the dehazing task, PhyDAE demonstrated excellent performance with a PSNR of 39.12 dB and an SSIM of 0.9928. While its LPIPS score (0.0227) was slightly higher than some methods, its overall performance remained state-of-the-art. Notably, the physics-constrained mechanism enabled PhyDAE to maintain stable restoration quality when handling the unique band-correlated haze present in Landsat imagery. For deblurring, PhyDAE obtained a PSNR of 36.88 dB and an SSIM of 0.8824. Although the IDR method showed a slight edge in SSIM, PhyDAE exhibited a better balance in structural preservation, with an LPIPS score of 0.1487 indicating high perceptual quality. For denoising, PhyDAE remained highly competitive with a PSNR of 34.53 dB and an SSIM of 0.8651. Its LPIPS score of 0.0991, in particular, proved its ability to preserve fine details while effectively suppressing noise. The low-light enhancement experiments underscored PhyDAE's strengths, as it achieved the best performance with a PSNR of 42.24 dB and an SSIM of 0.9949. Its LPIPS score (0.0121) was only marginally higher than some highly specialized methods, confirming its excellent overall restoration quality. The visualizations in Fig.~\ref{MDRS} further validate that PhyDAE produces results consistent with the physical laws of remote sensing imaging, demonstrating a superior balance in preserving both the spectral features and spatial textures of ground objects in Landsat data.

\subsubsection{Computational Efficiency Analysis}

\begin{table}[tbp]
	\centering
	\caption{Computational efficiency comparison in terms of parameters (M), memory consumption (MB), and FLOPs (G) for processing 224$\times$224 images}
	\label{tab:complexity}
	\begin{tabular}{@{}lccc@{}}
		\toprule
		Method & Params (M) & Memory (MB) & FLOPs (G) \\
		\midrule
		PromptIR   & 35.59 & 135.77 & 172.71 \\
		MoCEIR     & 25.35 &  96.72 & 104.03 \\
		Restormer  & 26.13 &  99.72 & 140.99 \\
		DGUNet     & 69.57 & 265.41 & 211.71 \\
		Uformer-AP & 21.24 &  81.09 &  91.12 \\
		\midrule
		\textbf{PhyDAE} & \textbf{17.21} & \textbf{65.66} & \textbf{71.63} \\
		\bottomrule
	\end{tabular}
\end{table}

In practical remote sensing applications, computational efficiency is a critical factor for deployment feasibility and utility. Table~\ref{tab:complexity} compares the computational complexity of the different all-in-one restoration methods on $224 \times 224$ images. The results show that PhyDAE holds a significant advantage, achieving the best efficiency among all baselines with only 17.21M parameters, 65.66MB of memory usage, and 71.63 GFLOPs. Specifically, PhyDAE reduces the parameter count by 19.0\% compared to the next smallest model, Uformer-AP (21.24M). It also reduces memory overhead by 19.0\% compared to Uformer-AP (81.09MB) and cuts floating-point operations by 21.4\% relative to the next best, Uformer-AP (91.12G). More strikingly, when compared to the mainstream method PromptIR, PhyDAE achieves massive reductions of 51.6\% in parameters, 51.6\% in memory usage, and 58.5\% in computational load, all while maintaining leading restoration performance.

PhyDAE's high efficiency is attributed to its meticulously designed lightweight architecture and physics-prior-guided mechanism. First, its two-stage cascaded structure decomposes the task, avoiding the parameter redundancy inherent in single, complex networks that handle multiple degradations. The first stage performs coarse restoration, while the second stage's expert network refines only the residual features, substantially reducing computational overhead. Second, the expert routing mechanism, based on Top-K sparse activation, ensures only a small subset of relevant experts is activated per forward pass, lowering the computational complexity from being proportional to the total number of experts to a constant level. Furthermore, the residual manifold projector compresses the feature space via low-dimensional representation learning, while the frequency-aware degradation decomposer leverages physical priors to guide feature extraction, preventing a blind learning process. This physics-constrained design not only enhances model interpretability but also significantly improves parameter efficiency by narrowing the search space. These results confirm that PhyDAE successfully achieves its design goal of optimizing computational efficiency without compromising restoration quality, providing a vital technical foundation for the practical deployment and large-scale application of remote sensing image restoration.

\subsection{Comparative Experimental Results}

\subsubsection{Core Component Ablation}

\begin{table*}[tbp]
	\centering
	\caption{Ablation study on core components of PhyDAE. Performance is reported as 4D-AVG PSNR (average PSNR across four degradation types) on three benchmark datasets}
	\label{tab:ablation}
	\small
	\begin{tabular}{@{}lcccc|ccc@{}}
		\toprule
		& \multicolumn{4}{c|}{Component Configuration} & \multicolumn{3}{c@{}}{Performance (4D-AVG PSNR $\uparrow$)} \\
		\cmidrule(lr){2-5} \cmidrule(l){6-8}
		Method & \makebox[0.8cm]{RMP} & \makebox[0.85cm]{FADD} & \makebox[1cm]{Physical} & \makebox[1.1cm]{Two-} & \makebox[1.1cm]{MD-} & \makebox[1.2cm]{MD-} & \makebox[1.2cm]{MDRS-} \\
		&     &      & \makebox[1cm]{Expert}   & \makebox[1.1cm]{stage}   & \makebox[1.1cm]{RSID} & \makebox[1.2cm]{RRSHID} & \makebox[1.2cm]{Landsat} \\
		\midrule
		Baseline            & $\times$ & $\times$ & $\times$ & $\times$ & 26.42 & 29.16 & 34.28 \\
		+RMP                & \checkmark & $\times$ & $\times$ & $\times$ & 27.70 & 30.29 & 35.55 \\
		+FADD               & $\times$ & \checkmark & $\times$ & $\times$ & 27.09 & 29.68 & 34.94 \\
		+Physical Expert    & $\times$ & $\times$ & \checkmark & $\times$ & 28.18 & 31.01 & 36.09 \\
		+Two-stage          & $\times$ & $\times$ & $\times$ & \checkmark & 27.86 & 30.57 & 35.82 \\
		+RMP+FADD           & \checkmark & \checkmark & $\times$ & $\times$ & 28.64 & 31.52 & 36.96 \\
		\midrule
		\textbf{PhyDAE}     & \checkmark & \checkmark & \checkmark & \checkmark & \textbf{29.83} & \textbf{32.30} & \textbf{38.19} \\
		\bottomrule
	\end{tabular}
\end{table*}
\begin{table*}[tbp]
	\centering
	\caption{Ablation analysis of loss function components on the MD-RSID dataset, demonstrating the contribution of each loss term to overall performance}
	\label{tab:ablation_loss}
	\begin{tabular}{@{}lcccccc@{}}
		\toprule
		\multirow{2}{*}{Method} & \multicolumn{3}{c}{Loss Configuration} & \multicolumn{3}{c}{Performance Metrics (4D-AVG)} \\
		\cmidrule(lr){2-4} \cmidrule(lr){5-7}
		& $\mathcal{L}_{\text{DAOT}}$ & $\mathcal{L}_{\text{balance}}$ & $\mathcal{L}_{\text{contrast}}$ & PSNR$\uparrow$ & SSIM$\uparrow$ & LPIPS$\downarrow$ \\
		\midrule
		Adaptive Pixel Loss & $\times$ & $\times$ & $\times$ & 27.84 & 0.8295 & 0.1947 \\
		+DAOT Loss         & $\checkmark$ & $\times$ & $\times$ & 28.62 & 0.8617 & 0.1722 \\
		+Expert Balance    & $\checkmark$ & $\checkmark$ & $\times$ & 29.07 & 0.8847 & 0.1654 \\
		+Contrastive Loss  & $\checkmark$ & $\checkmark$ & $\checkmark$ & \textbf{29.83} & \textbf{0.9026} & \textbf{0.1539} \\
		\bottomrule
	\end{tabular}
\end{table*}

To systematically evaluate the effectiveness of each core component in PhyDAE, we designed a progressive ablation study. The baseline model is a standard Transformer encoder-decoder architecture with all proposed components removed. As shown in Table~\ref{tab:ablation}, every core component of PhyDAE contributes significantly to the final performance, and a clear synergy exists among them.

Analysis of individual contributions reveals the independent value of each module. The Residual Manifold Projector (RMP) yielded PSNR gains of 1.28 dB, 1.13 dB, and 1.27 dB on the three datasets, respectively, demonstrating the efficacy of mapping transmission residuals onto a structured manifold. The Frequency-Aware Degradation Decomposer (FADD) achieved performance increases of 0.67 dB, 0.52 dB, and 0.66 dB through multi-scale frequency-domain feature extraction, highlighting the critical role of frequency priors in identifying degradation patterns. The Physics Expert Module provided the most substantial individual contribution, achieving significant gains of 1.76 dB, 1.85 dB, and 1.81 dB on the MD-RSID, MD-RRSHID, and MDRS-Landsat datasets, respectively. This result underscores the importance of explicitly modeling the physical mechanisms of degradation within the network architecture. The two-stage cascade mechanism, through an iterative residual refinement strategy, improved performance by 1.44 dB, 1.41 dB, and 1.54 dB, indicating that a hierarchical strategy---from coarse to fine-grained restoration---can effectively address the non-linear characteristics of complex degradations in remote sensing imagery.

Analysis of component synergy further confirms the validity of our architectural design. The combination of RMP and FADD produced a synergistic performance improvement, reaching 28.64 dB, 31.52 dB, and 36.96 dB on the three datasets. These results surpass the best-performing single component by 0.94 dB, 1.23 dB, and 1.41 dB, respectively. This synergy stems from the complementary nature of residual analysis and frequency-domain feature extraction for degradation characterization: RMP captures the geometric manifold structure of residuals, while FADD reveals their spectral fingerprints. Together, they provide richer and more precise prior information for the subsequent expert routing.

The complete PhyDAE model achieved state-of-the-art performance across all test datasets, outperforming the baseline by 3.41 dB, 3.14 dB, and 3.91 dB. This demonstrates the significant advantages of the fully integrated framework. Notably, the total performance gain exceeds the simple summation of individual component contributions, indicating a deep architectural synergy among the four modules. This non-linear enhancement manifests as follows: RMP provides multi-scale residual embeddings that supply task-specific conditional information to the physics experts; FADD generates frequency-domain features that guide the expert router toward more accurate degradation classification; and the two-stage mechanism creates an iterative optimization framework that allows each component to operate within a more refined feature space.

\subsubsection{Loss Function Ablation}

\begin{table*}[tbp]
	\centering
	\caption{Cross-dataset generalization results showing PSNR/SSIM values when models are trained on one dataset (rows) and tested on another (columns), with diagonal elements representing same-domain performance}
	\label{tab:cross_dataset}
	\begin{tabular}{@{}l|ccc@{}}
		\toprule
		\multirow{2}{*}{\makebox[2.5cm][l]{Training Set}} & \multicolumn{3}{c}{Test Set (PSNR/SSIM)} \\
		\cmidrule(lr){2-4}
		& MD-RSID & MD-RRSHID & MDRS-Landsat \\
		\midrule
		MD-RSID       & \textbf{29.83/0.9026} & 24.68/0.7446 & 26.64/0.8553 \\
		MD-RRSHID     & 23.58/0.7817 & \textbf{32.30}/0.8573 & 27.69/\textbf{0.8594} \\
		MDRS-Landsat  & 25.55/0.8730 & 25.16/0.7621 & \textbf{38.19/0.9338} \\
		\bottomrule
	\end{tabular}
\end{table*}

To investigate the contribution of each component within PhyDAE's multi-dimensional joint optimization strategy, we conducted a systematic ablation study of the loss functions on the MD-RSID dataset. As detailed in Table~\ref{tab:ablation_loss}, our experiment employed an incremental addition strategy, beginning with the adaptive pixel loss and progressively incorporating the Degradation-Aware Optimal Transport (DAOT) loss, the expert load-balancing loss, and the contrastive loss. This approach allowed for quantitative evaluation of each component's impact on model performance.

The results indicate that all loss components positively contribute to model performance and exhibit a strong synergistic effect. First, incorporating the DAOT loss improved PSNR by 0.78 dB and reduced LPIPS by 11.55\%, demonstrating the effectiveness of the Wasserstein distance-based distribution alignment in preserving geometric consistency. This loss term enhances the stability of the restoration process by establishing an optimal transport map from the degraded to the clean image distribution in the probability measure space. Furthermore, the load-balancing loss yielded a marginal PSNR gain of 0.45 dB. Its primary function is to ensure uniform activation of the expert networks, thereby preventing the model from collapsing into a single-expert system and improving its generalization to rare degradation types. The inclusion of the contrastive loss provided an additional 0.76 dB PSNR gain and, more importantly, lowered the LPIPS score to 0.1539. This improvement arises from the clustering constraint that contrastive learning imposes on the residual embedding space, which forces representations of similar degradations to be more compact while ensuring that representations of dissimilar degradations remain well-separated. Consequently, the model's degradation discrimination and expert routing accuracy are significantly enhanced. Notably, the full loss configuration enabled PhyDAE to achieve optimal image restoration performance, indicating that the multi-term loss function not only boosts pixel-level accuracy but also comprehensively improves structural fidelity and perceptual quality.

\subsubsection{Cross-Dataset Generalization}

To assess the generalization capability of PhyDAE, we trained models on each of the three datasets and performed cross-dataset evaluations. As shown in Table~\ref{tab:cross_dataset}, this experiment measures the model's transfer performance across different data domains. The diagonal elements represent in-domain results, whereas the off-diagonal elements reflect cross-domain generalization.

The results demonstrate that PhyDAE possesses robust cross-domain generalization capabilities. The model trained on MD-RSID achieved PSNR values of 24.68 dB and 26.64 dB when transferred to MD-RRSHID and MDRS-Landsat, corresponding to performance drops of 5.15 dB and 3.19 dB from its in-domain performance (29.83 dB). Similarly, the model trained on MD-RRSHID obtained PSNRs of 23.58 dB and 27.69 dB on MD-RSID and MDRS-Landsat, with performance decreases of 8.72 dB and 4.61 dB, respectively. The model trained on MDRS-Landsat exhibited the highest in-domain performance (38.19 dB) but experienced a more pronounced performance drop when transferred, falling to 25.55 dB on MD-RSID and 25.16 dB on MD-RRSHID (a decrease of 12.64 dB and 13.03 dB). This substantial decay is primarily because the MDRS-Landsat dataset is generated using physics-based modeling from Landsat 8 multi-band information. Its degradation patterns are more regular and its noise levels are lower. Consequently, the model trained on this data better fits regularized degradations but exhibits weaker generalization when confronted with the complex, real-world variations present in datasets like MD-RRSHID.

\begin{figure}[tbp]
	\centering
	
	\includegraphics[width=1.0\linewidth]{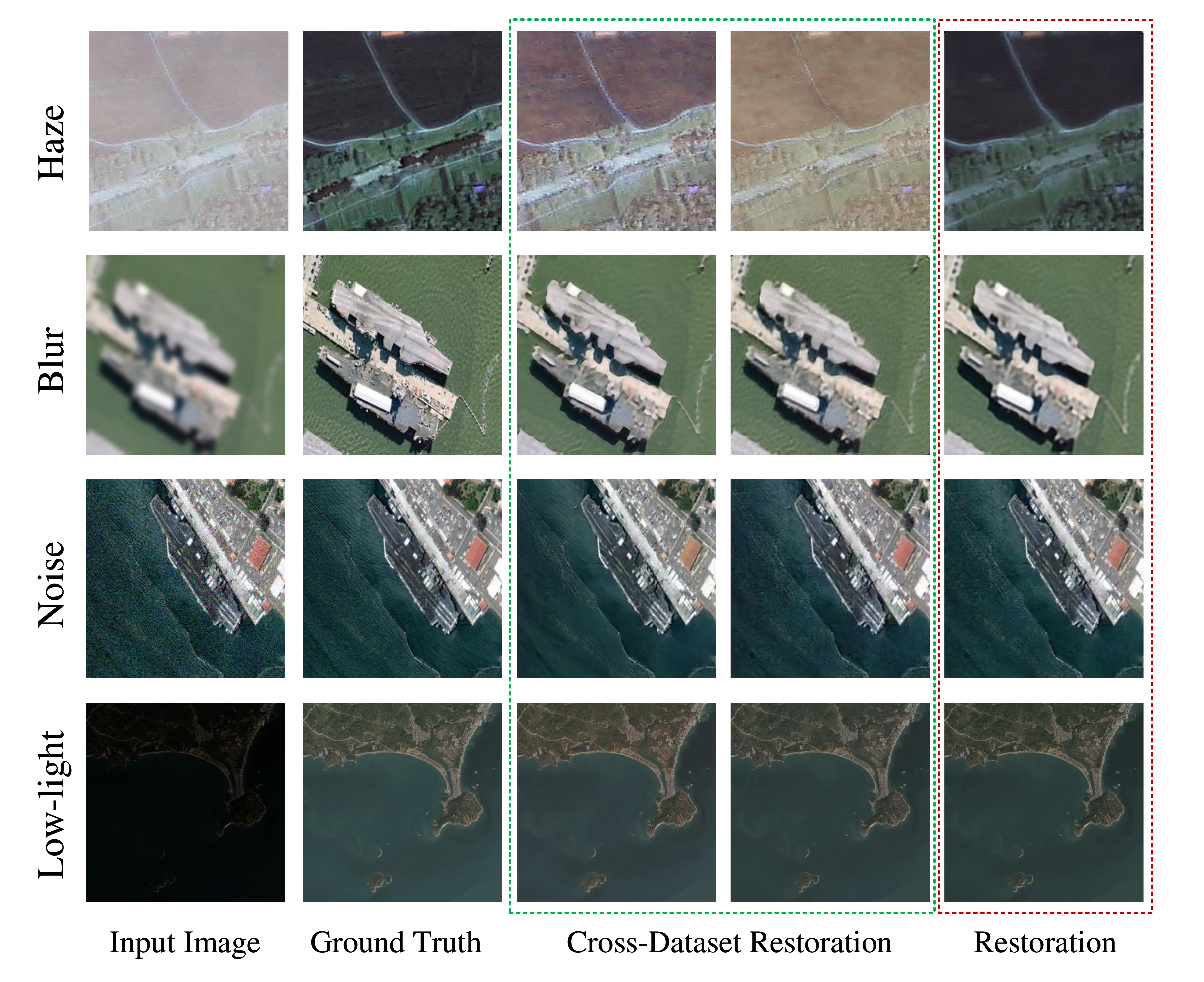}
	\caption{Cross-dataset generalization visualization results demonstrating the robustness of PhyDAE when trained on one dataset and tested on another, maintaining reasonable restoration quality despite domain shifts.}
	\label{cross_dataset}
\end{figure}

A comprehensive analysis shows that PhyDAE's average cross-domain PSNR is 25.55 dB, a drop of only 7.89 dB from the average in-domain performance of 33.44 dB, achieving a performance retention rate of 76.41\%. Furthermore, the visualization results in Fig.~\ref{cross_dataset} confirm that PhyDAE's mixed-degradation training and physics-constrained loss are effective at enhancing generalization. The RMP captures common geometric structures of degradations across domains by modeling them in a low-dimensional manifold. The multi-scale frequency features from FADD exhibit strong domain invariance. Finally, the imaging physics constraints embedded in the Physics Expert Module ensure restoration consistency across different data domains. These design elements work synergistically to enable reliable performance transfer, facilitating scene adaptation in practical applications.

\section{Conclusion}

This paper introduces the Physics-Prior-Guided Degradation-Adaptive Expert model (PhyDAE), a novel approach for integrated restoration of remote sensing images under complex degradation conditions. The proposed method employs a two-stage cascaded processing strategy with a progressive degradation mining mechanism, enabling the model to effectively capture multi-scale coupled degradation features. A Residual Manifold Projector and a Frequency-aware Degradation Decomposer work in synergy to perform deep modeling of degradation features, providing precise prior guidance for the selection of adaptive experts. By explicitly embedding physical constraints---such as the atmospheric scattering model, noise statistics, motion blur mechanism, and Retinex theory---the Physics-aware Expert Network ensures that restoration results adhere to physical imaging principles and significantly enhances model interpretability.

Extensive experiments validate the effectiveness and superiority of PhyDAE. On three benchmark datasets, MD-RSID, MD-RRSHID, and MDRS-Landsat, PhyDAE achieves state-of-the-art or competitive performance across four restoration tasks: dehazing, denoising, deblurring, and low-light enhancement. These results demonstrate its capability to deliver stable, high-quality restoration under challenging imaging conditions.

Notably, PhyDAE strikes an effective balance between high-fidelity restoration and computational efficiency. By leveraging a temperature-controlled Top-K sparse activation strategy and a lightweight architecture, PhyDAE achieves the best efficiency among comparable methods, requiring only 17.21M parameters and 71.63 GFLOPs. Compared to the mainstream method PromptIR, PhyDAE reduces the parameter count and computational cost by 51.6\% and 58.5\%, respectively. This excellent trade-off between performance and efficiency makes large-scale, high-quality remote sensing image restoration feasible in resource-constrained environments, holding significant value for real-world applications. Future work will focus on extending the applicability of PhyDAE to multi-modal remote sensing and more complex degradation scenarios. We will also investigate self-supervised learning strategies to reduce the reliance on paired training data and improve cross-domain generalization.

\bibliographystyle{elsarticle-num-names}
\bibliography{dzref}
\end{document}